\documentclass{article}

\usepackage{algorithm}
\usepackage{algpseudocode}
\usepackage{arxiv}
\usepackage{amsmath}
\usepackage[utf8]{inputenc} % allow utf-8 input
\usepackage[T1]{fontenc}    % use 8-bit T1 fonts
\usepackage{hyperref}       % hyperlinks
\usepackage{url}            % simple URL typesetting
\usepackage{booktabs}       % professional-quality tables
\usepackage{amsfonts}       % blackboard math symbols
\usepackage{nicefrac}       % compact symbols for 1/2, etc.
\usepackage{microtype}      % microtypography
\usepackage{graphicx}
\usepackage{natbib}
\usepackage{doi}
\usepackage{amssymb}
\usepackage{multirow}
\usepackage{tabularx}
\usepackage{tikz}
\usetikzlibrary{positioning}

\title{Model-agnostic Mitigation Strategies of Data Imbalance for Regression}

\date{\today}	% Here you can change the date presented in the paper title
\author{ \href{https://orcid.org/0000-0002-0728-0071}{\includegraphics[scale=0.06]{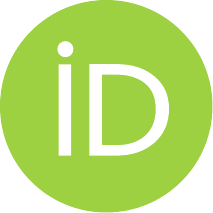}\hspace{1mm} Jelke Wibbeke}\thanks{Part of the work was done while affiliated with Energy Division, OFFIS - Institute of Information Technology, Oldenburg, Germany} \\
	Department of Computing Science\\
		Carl von Ossietzky Universit{\"a}t Oldenburg\\
	\texttt{jelke.wibbeke@uni-oldenburg.de}\\
	\And
    \href{https://orcid.org/0009-0004-6875-7912}{\includegraphics[scale=0.06]{orcid.pdf}\hspace{1mm}Sebastian Rohjans}\\
	Department for Civil Engineering,\\ Geoinformation and Health Technology\\
	Jade University of Applied Science\\
	\texttt{sebastian.rohjans@jade-hs.de}\\
	\And
	\href{https://orcid.org/0000-0002-1548-6547}{\includegraphics[scale=0.06]{orcid.pdf}\hspace{1mm}Andreas Rauh}\\
	Department of Computing Science\\
	Carl von Ossietzky Universit{\"a}t Oldenburg\\
	\texttt{andreas.rauh@uni-oldenburg.de}\\
}

\renewcommand{\shorttitle}{Model-agnostic Mitigation Strategies of Data Imbalance in Regression}

\hypersetup{
pdftitle={Model-agnostic Mitigation Strategies of Data Imbalance in Regression},
pdfsubject={Imbalanced Regression},
pdfauthor={Jelke Wibbeke, Sebastian Rohjans, Andreas Rauh},
pdfkeywords={Imbalanced regression, Supervised learning, Cost-sensitive learning, Sample weighting, Relevance function, Sampling methods, Model bias},
}

\begin{document}
\maketitle

\begin{abstract}
Data imbalance persists as a pervasive challenge in regression tasks, introducing bias in model performance and undermining predictive reliability. This is particularly detrimental in applications aimed at predicting rare events that fall outside of the domain of the bulk of the training data. In this study, we review the current state-of-the-art regarding sampling-based methods and cost-sensitive learning. Additionally, we propose novel approaches to mitigate model bias. To better assess the importance of data, we introduce the density-distance and density-ratio relevance functions, which effectively integrate empirical frequency of data with domain-specific preferences, offering enhanced interpretability for end-users. Furthermore, we present advanced mitigation techniques (cSMOGN and crbSMOGN), which build upon and improve existing sampling methods. In a quantitative evaluation, we benchmark state-of-the-art methods on 10 synthetic and 42 real-world datasets, using neural networks, XGBoosting trees and Random Forest models. Our analysis shows that while most strategies improve performance on rare samples, they degrade it on frequent ones. The trade-off becomes larger the more the performance on rare samples is increased. However, to reduce this effect we demonstrate that constructing an ensemble of models---one trained with imbalance mitigation and another without---can be used. The key findings underscore the superior performance of our novel crbSMOGN sampling technique with the density-ratio relevance function for neural networks, outperforming state-of-the-art methods.
\end{abstract}

% keywords can be removed
\keywords{Imbalanced regression \and Supervised learning \and  Cost-sensitive learning \and Sample weighting \and Relevance function \and Sampling methods \and Model bias}

\section{Introduction} \label{sec:intro}
Data imbalance remains a pervasive challenge in regression tasks, impacting the performance and reliability of predictive models, such as neural networks. Traditional loss functions, like mean squared error or mean absolute error, tend to favor majority classes, resulting in a biased model performance towards frequently occurring data points \citep{wibbeke2024quantification}. This bias is particularly detrimental in applications aimed at predicting rare events, which often fall outside of the domain covered by the bulk of the training data. Examples include predicting extreme weather conditions, building energy consumption, or fault detection in industrial systems~\citep{steininger2021density, zhang2021problem, ramentol2016fuzzy}.

While data imbalance has been extensively studied in classification problems, regression-specific strategies have only gained attention in the past years. The field still lacks a coherent framework for measuring and mitigating imbalance in continuous target variables. 

Several definitions of imbalance in regression exist in the literature. In this work, we adopt a formalization involving two distributions: the empirical distribution of the target variable and the domain-priority distribution, which encodes the practical importance of different target regions. Under this definition, imbalance arises when these distributions differ, indicating that the frequency of observed samples does not match their importance for the prediction task. This framework, originally proposed by \citet{kowatsch2024imbalance} and further developed by \citet{wibbeke2024quantification}, provides an interpretable and generalizable measure of imbalance without relying on arbitrary thresholds.

To formalize and operationalize the notion of importance, we introduce relevance functions. A relevance function maps the target space to a continuous scale of importance, where higher values correspond to regions that are more critical for the prediction task. This concept is fundamental: it defines which prediction errors are most consequential and thus informs both the design of mitigation strategies and, in many approaches, the construction of evaluation metrics. 

Evaluation metrics in imbalanced regression must reflect this notion of importance without introducing confounding dependencies. Some metrics directly incorporate relevance to prioritize errors in high-importance regions, while others, such as partitioned loss metrics, operate independently of relevance by discretizing the target space. By explicitly separating relevance from evaluation, a circular dependency is prevented where improvements in performance could be attributed either to the metric or to a specific choice of relevance.

Mitigation methods for imbalanced regression can be categorized into data-level and algorithm-level methods \citep{sadouk2021novel, krawczyk2016learning}. Data-level methods modify the training data, for example through over- or under-sampling \citep{torgo2013smote, branco2017smogn, wibbeke2022optimal}, whereas algorithm-level methods adjust the learning process, for instance via cost-sensitive learning \citep{steininger2021density, sadouk2021novel} or architecture modifications \citep{yang2021delving}. Relevance functions are used within these methods to weight samples according to their importance, ensuring that rare but critical observations are appropriately emphasized during training.

To avoid ambiguity in terminology, we use the following definitions. A relevance function maps samples to continuous relevance (importance) scores; a mitigation method is a procedure that uses these relevance scores to modify the training process; and a mitigation strategy refers to a specific combination of relevance function and mitigation method. Our full taxonomy is illustrated in Fig.~\ref{fig:taxonomy}.

\begin{figure}[ht]
    \centering
    \begin{tikzpicture}[
        def/.style={
            draw,
            rounded corners,
            text width=0.33\textwidth,
            align=left,
            inner sep=8pt
        },
        struc/.style={
            draw,
            rounded corners,
            text width=0.5\textwidth,
            align=left,
            inner sep=8pt
        },
        node distance=0.3cm and 0.2cm
    ]
    % Left column - one large box
    \node[struc] (structure) {
        \textbf{Article Structure:}\\
        \ref{sec:intro} \nameref{sec:intro} \\
        \ref{sec:related_work} \nameref{sec:related_work}\\  
        \ref{sec:methods} \nameref{sec:methods}\\
            \quad\ref{sec:relevance_function} \nameref{sec:relevance_function}\\  
                \qquad\ref{sec:pchip} \nameref{sec:pchip}\\
                \qquad\ref{sec:histogram-based_relevance} \nameref{sec:histogram-based_relevance}\\ 
                \qquad\ref{sec:label_distribution_smoothing} \nameref{sec:label_distribution_smoothing}\\ 
                \qquad\ref{sec:kernel_density_relevance} \nameref{sec:kernel_density_relevance}\\  
                \qquad\ref{sec:denseweight} \nameref{sec:denseweight}\\  
                \qquad\ref{sec:density_dist_relevance} \nameref{sec:density_dist_relevance}\\  
                \qquad\ref{sec:density_ratio_relevance} \nameref{sec:density_ratio_relevance}\\
            \quad\ref{sec:mitigation_methods} \nameref{sec:mitigation_methods}\\  
                \qquad\ref{sec:smoter} \nameref{sec:smoter}\\  
                \qquad\ref{sec:smogn} \nameref{sec:smogn}\\  
                \qquad\ref{sec:wercs} \nameref{sec:wercs}\\  
                \qquad\ref{sec:wsmoter} \nameref{sec:wsmoter}\\  
                \qquad\ref{sec:denseloss} \nameref{sec:denseloss}\\
                \qquad\ref{sec:sera} \nameref{sec:sera}\\ 
                \qquad\ref{sec:probloss} \nameref{sec:probloss}\\ 
                \qquad\ref{sec:bmc} \nameref{sec:bmc}\\  
                \qquad\ref{sec:csmogn} \nameref{sec:csmogn}\\  
                \qquad\ref{sec:crbsmogn} \nameref{sec:crbsmogn}\\  
            \quad\ref{sec:evaluation_metrics} \nameref{sec:evaluation_metrics}\\
                \qquad\ref{sec:diagram_based_eval} \nameref{sec:diagram_based_eval}\\
                \qquad\ref{sec:precision_and_recall} \nameref{sec:precision_and_recall}\\
                \qquad\ref{sec:loss_from_cost_sensitive} \nameref{sec:loss_from_cost_sensitive}\\
                \qquad\ref{sec:partitioned_loss_metrics} \nameref{sec:partitioned_loss_metrics}\\
            \quad\ref{sec:ensemble_formation} \nameref{sec:ensemble_formation}\\  
        \ref{sec:experiment} \nameref{sec:experiment}\\  
            \quad\ref{sec:impact-of-sampling} \nameref{sec:impact-of-sampling} \\
                \qquad\ref{sec:dual-source-relevance-functions} \nameref{sec:dual-source-relevance-functions}\\
                \qquad\ref{sec:comparison-of-mitigation-methods} \nameref{sec:comparison-of-mitigation-methods}\\
            \quad\ref{sec:performance_evaluation} \nameref{sec:performance_evaluation}\\  
                \qquad\ref{sec:case_study_synthetic} \nameref{sec:case_study_synthetic}\\  
                \qquad\ref{sec:case_study_real_world_data} \nameref{sec:case_study_real_world_data}\\  
                \qquad\ref{sec:case_study_ensembles} \nameref{sec:case_study_ensembles}\\  
        \ref{sec:discussion} \nameref{sec:discussion}\\  
        \ref{sec:outlook} \nameref{sec:outlook}\\  
        \ref{sec:conclusion} \nameref{sec:conclusion}
    };   
    % Right column - two stacked boxes
    \node[def, right=of structure.north east, anchor=north west] (miti_strate) {
        \textbf{Mitigation strategy:}\\
        A \emph{mitigation strategy} refers to the combination of \emph{relevance function} and \emph{mitigation method} to reduce the model bias caused by data imbalance.
    };
    \node[def, below=of miti_strate] (miti_method) {
        \textbf{Mitigation method:}\\
        A \emph{mitigation method} is a method that uses the relevance of each sample obtained by a \emph{relevance function} as a weight in an arbitrary approach to mitigate the model bias caused by data imbalance.
    };
    \node[def, below = of miti_method] (rel_func) {
        \textbf{Relevance function:}\\
        A \emph{relevance function} is a continuous function that expresses an application-specific bias in mapping the \emph{empirical distribution} and/or \emph{domain-priority distribution} into a scale of relevance, where higher values indicate higher relevance of the sample.
    };
    \node[def, below= of rel_func] (empirical) {
        \textbf{Empirical distribution:}\\
        The \emph{empirical distribution} refers to the probability distribution of the data.
    };
    \node[def, below= of empirical] (rel_dist) {
        \textbf{Domain-priority distribution:}\\
        The \emph{domain-priority distribution} refers to the probability distribution associated with the application-specific domain preference/importance.
    };
    \end{tikzpicture}
    \caption{Structure and taxonomy of the article.} \label{fig:taxonomy}
\end{figure}

However, currently, no relevance function utilizes the difference-based interpretation of imbalance including both the empirical and domain-priority distribution. Furthermore, no comprehensive comparison of existing mitigation strategies has been conducted. 

The central research question addressed in this paper is: How can model-agnostic strategies, including novel methods that utilize both the empirical and domain-priority distribution, improve predictive performance in regression tasks under data imbalance?

To answer this question, we propose new relevance functions and mitigation methods. In a benchmark, we compare the performance of established mitigation strategies. Furthermore, we investigate the benefits of ensemble strategies to mitigate potential negative side effects of imbalance mitigation.

The main contributions of this paper are the following:

\begin{itemize}
    \item We propose two novel relevance functions to assess the importance of the data samples.
    \item We introduce two novel model-agnostic mitigation methods based on sampling to reduce the model bias caused by data imbalance in regression tasks.
    \item We compare the performance of different combinations of relevance functions and mitigation methods in a quantitative benchmark.
    \item We demonstrate that constructing an ensemble of two models can reduce the negative effects of imbalance mitigation strategies on model performance.
\end{itemize}

We focus on model-agnostic mitigation strategies for regression tasks. These include mostly data-level methods, while making an exception for algorithm-based methods by including cost-sensitive learning. This exception is justified by the model type independence of cost-sensitive learning, unlike other algorithm-based methods that are model-specific (e.g., additional output layers in neural networks, which cannot be applied to regression trees).

The structure of the article is shown in Fig \ref{fig:taxonomy} and the remainder of the article is outlined as follows: In Sec. \ref{sec:related_work}, the related work and state-of-the-art are presented. In Sec. \ref{sec:methods}, relevance functions and mitigation methods and evaluation metrics are presented including new methods based on the most recent definition of data imbalance and ensemble-based approaches. In Sec. \ref{sec:experiment}, various combinations of relevance functions and mitigation methods are evaluated and compared. First, the influence of sampling-based mitigation strategies on the target variable distribution is shown. Second, the performance of mitigation strategies is assessed in a quantitative benchmark on two case studies using synthetic and real-world data. Thirdly, the proposed ensembles are analyzed. Following the evaluation, the results are discussed in Sec. \ref{sec:discussion}. Possible subsequent applications and research work are presented in the outlook in Sec. \ref{sec:outlook}. Lastly, the results and takeaways of the presented approach and analysis are concluded in Sec. \ref{sec:conclusion}. The appendix contains material about the used datasets and model hyperparameters.

\section{Related work}\label{sec:related_work}
Many studies published to date have focused on imbalanced classification. However, few address imbalanced regression \citep{branco2016survey}, albeit it has been demonstrated that classification and regression cannot be treated analogously \citep{yang2021delving,kowatsch2024imbalance}. 

Regarding the definition of imbalance, \citet{branco2016survey} and \citet{ribeiro2011utility} describe imbalance as a situation in which the user assigns higher importance to the performance on a subset of the range of the target variable $\mathbf{Y}$. This user preference bias is expressed by a relevance function $\phi(\mathbf{Y}): \mathbb{R}\rightarrow[0,1]$, mapping the target variable to a continuous importance scale between 0 and 1, where 1 denotes maximal importance. Imbalance is present if $\phi (\mathbf{Y})$ is non-uniform. While this formulation enables the assignment of a relevance value to each sample, quantifying the overall magnitude of imbalance requires the specification of a relevance threshold to separate low- and high-relevance subsets.

An alternative definition is provided by \citet{kowatsch2024imbalance}, who define balance explicitly and characterize imbalance as a deviation from it. In this framework, data imbalance is present if the probability measure of the empirical data deviates from a relevance measure. As the two respective measures, the probability density function of the empirical data and the probability density function associated with the relevance of the target domain are used. To quantify the imbalance, the Wasserstein metric is proposed, which is related to the distance between the two probability density functions. This definition was taken up by \citet{wibbeke2024quantification}, who propose new metrics to quantify the imbalance in data, suggesting the use of the ratio of the two distributions instead of the Wasserstein metric to provide better generalization and improved interpretability.
In this work, we adopt the same conceptual framework to define imbalance. However, to avoid terminological ambiguity, we refer to the relevance distribution of the target domain as the domain-priority distribution, as the term "relevance" is otherwise used in multiple contexts. This distinction ensures clarity while preserving the theoretical foundations of the original definition.

As elaborated by \citet{krawczyk2016learning} and \citet{sadouk2021novel}, mitigation strategies can broadly be divided into data-level and algorithm-level approaches. Data-level methods alter the training distribution, for instance via over- and/or under-sampling \citep{torgo2013smote, branco2017smogn, wibbeke2022optimal, branco2019pre}. In general, sampling-based methods rely on random sampling, interpolation between neighboring samples, synthetic generation through additive noise, or combinations thereof \citep{branco2019pre, avelino2024resampling}. Algorithm-level methods instead modify the learning procedure directly, e.g., by architectural adjustments \citep{yang2021delving} or by cost-sensitive learning, where the loss function is reweighted according to sample relevance \citep{steininger2021density, sadouk2021novel}. A detailed overview of the considered mitigation techniques is provided in Sec.~\ref{sec:methods}.

Beyond classical interpolation- and noise-based synthetic generation, first deep learning (DL)-based approaches for sample generation in imbalanced regression have emerged, such as autoencoders \citep{liu_research_2024, belhaouari_oversampling_2024}. In \citep{liu_research_2024}, generative models are employed to modify the target distribution; however, the study focuses on distributional changes and does not provide a systematic evaluation of downstream predictive performance. In contrast, \citep{belhaouari_oversampling_2024} integrates deep learning into the prediction of the target variable, while feature generation relies on k-nearest neighbors, thus not constituting a fully generative DL-based resampling strategy. 

A fundamental limitation of DL-based sample generation lies in the fact that the generative model itself is subject to bias. The synthetic samples may therefore be biased, and when these biased samples are combined with minority data for training, the resulting predictor can be influenced by a double-bias effect: one arising from the original data imbalance and another from the biased generated samples. Moreover, both \citep{liu_research_2024} and \citep{belhaouari_oversampling_2024} explicitly restrict their approaches to very large datasets to ensure stable training of the deep models. Given the substantial data requirements and the risk of propagating model-specific inductive biases into the augmented dataset, we exclude DL-based generation methods from our empirical evaluation and focus on model-agnostic resampling and weighting techniques.

Independently of the specific mitigation strategy, most approaches rely on some form of importance, rarity, or relevance score assigned to each data point to determine whether it should be weighted, replicated, or removed. The construction of such relevance functions constitutes a key difference between imbalanced classification and regression. In classification, minority and majority classes are typically known or directly inferred from class frequencies, as in the synthetic minority over-sampling technique (SMOTE) by \citet{chawla2002smote}. For continuous targets, however, rarity and relevance are not inherently discrete, raising the question of how to quantify the importance of individual samples.

\citet{torgo2007utility} propose to define relevance inversely proportional to sample rarity, linking it to the probability density function of the target variable. They discretize the target into quartiles and compute relevance via a sigmoid function parametrized by the interquartile range. This approach is, however, tailored to distributions with approximately normal shape and pronounced tails.

\citet{ribeiro2011utility} extend this idea by using boxplot statistics to define control points and interpolating between them with piecewise cubic Hermite polynomials, resulting in smoother relevance transitions. This formulation has been adopted by subsequent works~\citep{branco2017smogn, branco2019pre, ribeiro2020imbalanced, silva2022model}. Additionally, \citet{ribeiro2011utility} allow user-defined control points, increasing flexibility in specifying domain-specific relevance structures.

Adding to the notion that the relevance of a sample is inversely proportional to its rarity, \citet{aminian2025histogramapproachesimbalanceddata} and \citet{yang2021delving} propose to calculate the relevance based on the normalized inverse of the target value distribution discretized by histogram bins, where \citet{yang2021delving} smooth the bins beforehand using a symmetric kernel.

Further approaches also operationalize relevance as an inverse measure of rarity. \citet{aminian2025histogramapproachesimbalanceddata} and \citet{yang2021delving} compute relevance from histogram-based estimates of the target distribution, with \citet{yang2021delving} applying kernel smoothing to the discretized bins. Kernel density estimation (KDE)-based relevance functions are proposed by \citet{steininger2021density, sadouk2021novel, camacho2024wsmoter, yang2021delving}. These density estimates are then used either in cost-sensitive learning frameworks \citep{steininger2021density, sadouk2021novel, yang2021delving} or for relevance-guided over-sampling \citep{camacho2024wsmoter}. Additionally, \citet{yang2021delving} introduce feature-space distribution smoothing through a calibration layer integrated into the predictive model.

Despite the diversity of relevance formulations and mitigation methods, existing approaches do not incorporate the discrepancy between the empirical target distribution and a domain-specific priority distribution as proposed by \citet{kowatsch2024imbalance} and \citet{wibbeke2024quantification} into the construction of relevance mappings. Moreover, a comprehensive comparison of relevance functions in combination with different mitigation methods is still lacking. Initial comparative analyzes are presented in \citep{avelino2024resampling}; however, the influence of the chosen relevance function is not systematically examined. Therefore, an extensive and structured overview of the most widely used relevance functions, mitigation methods and evaluation metrics is provided in the following section.

\section{Methods} \label{sec:methods}
In this section, we present relevance functions, model-agnostic mitigation strategies and metrics to evaluate the biasing effect of data imbalance on regression performance. Without aiming to be exhaustive, we cover the most commonly used approaches in current research.

A fundamental concept underlying many approaches in imbalanced regression is the notion of relevance. Relevance functions assign an importance value to each observation in the training data and thereby formalize which samples are considered critical. This notion determines how imbalance is quantified and directly influences both the design of mitigation strategies and the construction of (most) evaluation metrics. We therefore begin by introducing relevance functions and discussing their formal properties and practical implications in Section~\ref{sec:relevance_function}.

Based on an assumed notion of relevance, mitigation methods aim to reduce the bias of predictive models toward frequent target regions. These methods operate at data or algorithmic level and typically use relevance information to reweight, resample, or otherwise adapt the training process to improve performance in under-represented yet important regions of the target space. A selection of mitigation methods is presented in Section~\ref{sec:mitigation_methods}.

Finally, in section~\ref{sec:evaluation_metrics}, we discuss evaluation metrics for imbalanced regression. Performance assessment in this setting is non-trivial, as standard regression metrics are dominated by frequent observations.

Regarding the mathematical notation, we consider a dataset consisting of $N$ samples with corresponding target values and features. We denote the dataset as $\mathbf{D}=(\mathbf{Y}, \mathbf{X})$, where $\mathbf{Y} \in \mathbb{R}^N$ is the target vector $\mathbf{Y} = [y_1, y_2, \cdots, y_N]$, containing the target value $y_i$ for each sample $i = 1, \dots, N$. Similarly, the model $M$ approximating the target variable is denoted as $\mathbf{\hat{Y}} = M(\mathbf{X})= [\hat{y}_1, \hat{y}_2, \cdots, \hat{y}_N]$. The feature matrix $\mathbf{X} \in \mathbb{R}^{N \times d}$ contains the $d$-dimensional feature vectors $\mathbf{x}_i$ for each sample. Each individual sample is thus represented by a pair $(y_i, \mathbf{x}_i)$, where $y_i \in \mathbb{R}$ and $\mathbf{x}_i \in \mathbb{R}^d$. This notation will be used throughout the following sections to refer to the dataset structure.

\subsection{Relevance functions}\label{sec:relevance_function}
A relevance function assigns an importance value to each sample based on arbitrary information associated with that sample. This information may stem from domain knowledge, empirical properties of the dataset, or any other modeling-related criteria. While in many practical settings relevance is derived from the empirical target distribution, this is not a mandatory requirement.

Formally, we introduce a sample descriptor $\mathbf{z}_i = g(y_i, \mathbf{x}_i; \mathbf{D})$, where $g(\cdot)$ is an application-dependent mapping that extracts arbitrary information related to sample $i$. A relevance function is defined as a continuous mapping $\phi: \mathcal{Z} \rightarrow \mathbb{R}$, where $\mathcal{Z}$ denotes the domain of admissible sample descriptors $\mathbf{z}$. The resulting relevance value of sample $i$ is given by $w_i = \phi(\mathbf{z}_i)$. Higher values of $w_i$ indicate higher relevance of the sample. 

In practice, a common special case is $\mathbf{z}_i = y_i$, yielding $w_i = \phi(y_i)$, where relevance is solely determined by the target variable, often based on its empirical distribution. However, the proposed definition allows for arbitrary dependencies beyond the target variable.
 
Note that our definition deviates from the established one proposed by \citet{torgo2007utility}, which restricts the value range of the relevance to between 0 and 1, i.e., $\phi(\mathbf{Y}): \Re\rightarrow[0,1]$.

In the following, we present and discuss several relevance functions. These include:
\begin{itemize}
    \item \nameref{sec:pchip}, \nameref{sec:histogram-based_relevance}, \nameref{sec:label_distribution_smoothing}, \nameref{sec:kernel_density_relevance} and \nameref{sec:denseweight}.
\end{itemize}

All of these assume that relevance can be quantified by a single distribution, whether being it the distribution of the empirical data itself or an arbitrary distribution associated with importance of the target domain. We therefore classify such relevance functions as single-source relevance functions.

Based on the premise set out in \citep{wibbeke2024quantification} and \citep{kowatsch2024imbalance}, which suggests that data imbalance should better be quantified by two distributions (one representing the empirical distribution of the data and the other a measure of domain priority), two new methods are introduced, transferring the concept to relevance functions:

\begin{itemize}
    \item \nameref{sec:density_dist_relevance} and \nameref{sec:density_ratio_relevance}.
\end{itemize}
As these new functions include both measures, we classify them as dual-source relevance functions.

\subsubsection{Interpolation with control points} \label{sec:pchip}
To derive a function inversely proportional to the target density \citet{torgo2007utility} proposed a sigmoid-like function to establish a smooth relevance function using the statistics provided by boxplots that summarize the distribution of the target variable as control points. However, this limited the approach to distributions of normal-like shape with very marked tails.

The approach was taken on by \citet{ribeiro2011utility}, who proposed to interpolate between the control points using piecewise cubic Hermite interpolating polynomials (pchip). The resulting relevance function goes through the set of $n$ provided points, giving the user more freedom to shape the function, as user-defined control points can be added. It is assumed that the control points refer to local extrema of the relevance, have a derivative of zero and are limited between 0 and 1. As a non-parametric approach to obtain the control points \citet{ribeiro2011utility} propose the aforementioned boxplot statistics. This way, the relevance is inversely proportional to the frequency of the sample. However, using the boxplot statistics limits the approach to normal-like or uni-modal distributions.

\subsubsection{Histogram-based relevance} \label{sec:histogram-based_relevance}
The histogram-based relevance approach was originally proposed for data stream processing, with a particular emphasis on low computational complexity \citep{aminian2025histogramapproachesimbalanceddata}. It determines the relevance of individual data points based on the discretized distribution of the target variable, as obtained through histogram binning.

First, an equal-width histogram of the target variable is constructed using $k$ bins. The "density" of each bin is evaluated by calculating the relative frequency of samples in each bin, denoted as $p_j$. This is done by normalizing the sample count in each bin with respect to the maximum bin count across all bins according to

\begin{equation}
p_j = \frac{b_j}{\text{max}\{b_l : l = 1,\ldots, k\}},
\end{equation}

where $b_j$ and $b_l$ represent the number of samples in the $j$-th and $l$-th bin, respectively. The relevance of a sample with target value $y_i$ is computed as

\begin{equation}
\phi(y_i) = 1 - p_j,
\end{equation}
where $p_j$ corresponds to the relative sample count of the bin to which $y_i$ belongs.

\subsubsection{Label distribution smoothing} \label{sec:label_distribution_smoothing}
Label Distribution Smoothing also assumes that the relevance of a sample is dependent on the distribution of the target variable. However, \citet{yang2021delving} argue that a smoothed distribution of the target variable should be considered, since data-driven models may extrapolate from frequently occurring samples to nearby rare ones. Therefore, they propose estimating the effective label density distribution $\tilde{p}(y_i)$ by convolving the empirical density $p(y)$ with a Gaussian kernel $k(y, y')$ according to
\begin{equation}
\tilde{p}(y_i) = \int k(y, y_i)p(y)dy.
\end{equation}
Here, the empirical density $p(y)$ is approximated using a histogram, with the density for $y_i$ determined by the number of occurrences in the corresponding histogram bin. To compute the relevance, the effective density is inverted as \begin{equation}
\phi(y_i) = \frac{1}{\tilde{p}(y_i)}.
\end{equation}
However, this inversion causes weights to approach infinity for very small densities, which is particularly problematic for out-of-distribution samples and makes the relevance function highly sensitive to extreme distortions. In the work by \citet{yang2021delving}, this issue is addressed on a per-dataset basis by clipping the effective density $\tilde{p}(y)$ to avoid extreme values. Nonetheless, the resulting relevance function is not bounded between 0 and 1, which limits its applicability in many mitigation methods.

To ensure compatibility, it is possible to first scale the effective density to the range [0, 1] and invert it according to
\begin{equation}
\phi(y_i) = \max(1 - p_{\text{scaled}}(y_i), \epsilon),
\end{equation}
where $p_{\text{scaled}}$ denotes the scaled effective density, and $\epsilon$ is a small constant introduced to avoid zero weights.

\subsubsection{Kernel density relevance} \label{sec:kernel_density_relevance}
Kernel density relevance makes use of the assumption, that the relevance of data is inversely proportional to the target variable probability distribution. To estimate the distribution, kernel density estimation (KDE) is used \citep{sadouk2021novel}.

To ensure general applicability independent of the respective dataset the density values are normalized between 0 and 1. Subsequently, the normalized density is inverted, such that low-density regions, typically associated with rare or extreme values in the target variable, are assigned higher relevance, while high-density regions, where the target values are more common, receive lower relevance. 

\subsubsection{DenseWeight} \label{sec:denseweight}
Complementary to kernel density relevance, DenseWeight approximates the density of the dataset using KDE \citep{steininger2021density}. However, in addition to normalizing the density between 0 and 1 and inverting the normalized density, a scaling factor $\alpha$ is added according to 
\begin{equation}
    \phi(\alpha, y_i) = \max(1 - \alpha p'(y_i),\epsilon),
\end{equation}
where $p'$ is the normalized density function and $\epsilon$ a small positive constant to prevent a weight from being zero; $\alpha$ is a hyper-parameter with which the impact of the density can be calibrated depending on the modeling task at hand. However, in their evaluation \citet{steininger2021density} find evidence for $\alpha=1$ being the most effective value, which allows it to be neglected.

In contrast to kernel density relevance, after inversion the weights are normalized by the mean of all weights to be independent from the respective learning algorithms' learning rate.

\subsubsection{Density-distance relevance} \label{sec:density_dist_relevance}
Density-distance relevance is a new dual-source relevance function, which is a generalization of DenseWeight and Kernel density relevance. It is based on the assumption that the relevance $w_i = \phi(y_i)$ of the $i$-th sample $y_i$ results from the distance $\Lambda_\text{dist}$ between the empirical probability density function of the target variable $f_\text{x}(\mathbf{Y})$ and the probability density function associated with the domain priority of the target variable $f_\text{p}(\mathbf{Y})$, according to
\begin{equation}
    \Lambda_\text{dist}(y_i) = f_\text{x}(y_i) - f_\text{p}(y_i).
\end{equation}
In this way, both a non-uniform distribution of the data and a non-uniform priority of the domain can be taken into account. Theoretically, the concept can also be extended to include the feature domain into the density functions \citep{wibbeke2024quantification}.

The density-distance $\Lambda_\text{dist}$ can take values within the interval $(-\infty, \infty)$ centered around zero. This means that $y_i$ is under-represented if $\Lambda_\text{dist}(y_i) < 0$ and over-represented if $\Lambda_\text{dist}(y_i) > 0$. To align the density-distance relevance with other relevance functions it is subsequently normalized and inverted, according to
\begin{align}
\Lambda'_\text{dist}(y_i) &= 
\begin{cases} 
0.5 - 0.5 \cdot \frac{\Lambda_\text{dist}(y_i)}{\min\left(\Lambda_\text{dist}\left(\mathbf{Y}\right)\right)} & \text{if } \Lambda_\text{dist}(y_i) < 0 \\
0.5 + 0.5 \cdot \frac{\Lambda_\text{dist}(y_i)}{\max\left(\Lambda_\text{dist}\left(\mathbf{Y}\right)\right)} & \text{if } \Lambda_\text{dist}(y_i) \geq 0
\end{cases},\\
    \phi(y_i) &= \max\left(1- \Lambda'_\text{dist}(y_i), \epsilon\right),
\end{align}
where $\epsilon$ is a small clipping value to avoid values of zero. The relevance is thus bounded within in the interval $[\epsilon, 1]$. The normalization centered around the point 0.5 preserves the knowledge at which point the probability density function of the target variable $f_\text{x}$ and the probability density function associated with the domain priority $f_\text{p}$ are balanced. At the same time, individual samples with $\Lambda_\text{dist}(y_i) \ll 0 \ll \Lambda_\text{dist}(y_i)$ do not cause the center of the relevance function to shift.

If uniform domain priority is considered, $f_\text{p}(y_i)$ is constant over all $i$ samples. The density-distance relevance is then similar to the kernel density relevance or DenseWeight (without normalization by the mean), apart from the fixed midpoint value. Density-distance relevance can thus be seen as an generalization of the aforementioned approaches, granting more flexibility by including empirical sample density and domain priority and by proving the user with a fixed midpoint value for better interpretation.

\subsubsection{Density-ratio relevance} \label{sec:density_ratio_relevance}
Density-ratio relevance is based on the assumption that the ratio of the values of two probability density functions at different points $y_i$ yields the relative probability of occurrence of an event. The probability ratio $\Lambda_\text{ratio}(y_i)$, according to
\begin{equation}
    \Lambda_\text{ratio}(y_i) = \frac{f_\text{x}(y_i)}{f_\text{p}(y_i)},
\end{equation}
directly expresses how often the sample $y_i$ occurs compared to how often it should occur with respect to its domain priority. Here, $f_\text{x}(\mathbf{Y)}$ denotes the probability density function of the empirical distribution of the target variable $\mathbf{Y}$ and $f_\text{p}(\mathbf{Y})$ denotes the probability density function associated with the priority of the domain of the target variable. The concept can also be extended to the feature space, as in \citep{wibbeke2024quantification}.

$\Lambda_\text{ratio} $ can take values within the interval $(0, \infty)$, where $\Lambda_\text{ratio}(y_i) = 1$ denotes that the empirical distribution of the data matches the domain-priority distribution, indicating that the sample is as frequent as required. If $\Lambda_\text{ratio}(y_i) <1$ the sample $y_i$ is under-represented, with $\Lambda_\text{ratio}(y_i) = 0.5$ indicating that the sample occurs half as often as assumed. The sample is over-represented if $\Lambda_\text{ratio}(y_i) > 1$, with $\Lambda_\text{ratio}(y_i) = 2$ indicating that the sample occurs twice as often as assumed.

However, as a relevance function, a larger value should indicate higher importance of the sample. Thus, to obtain the relevance of $\mathbf{Y}$, we use the inverted ratio, given by
\begin{equation}
    w_i = \phi(y_i) = \frac{1}{\Lambda_\text{ratio}(y_i)}.
\end{equation}
Now $w_i = 0.5$ indicates that the relevance of the sample is half as high as the optimal/nominal relevance of a sample, which is $w_i=1$.

Using the ratio makes the density-ratio relevance function non-symmetric and not bound between 0 and 1. It therefore requires dedicated mitigation methods to be used in. However, the density-ratio yields the advantage that $w_i$ directly expresses how severe the under-/over-representation of sample $y_i$ is and thus offers improved interpretability.

\subsection{Mitigation methods}\label{sec:mitigation_methods}
In the following, we discuss model-agnostic methods to mitigate the effects of data imbalance during model training. Besides sampling methods, we include cost-sensitive learning methods that only alter the loss function into said category, as altering the loss function does not influence the model architecture and loss functions can be chosen freely for many model types. The presented methods include:
\begin{itemize}
\item \nameref{sec:smoter}, \nameref{sec:smogn}, \nameref{sec:wercs}, \nameref{sec:wsmoter}, \nameref{sec:denseloss}, \nameref{sec:sera}, \nameref{sec:probloss} and \nameref{sec:bmc}.
\end{itemize}

Additionally, we propose two new sampling-based mitigation strategies:

\begin{itemize}
\item \nameref{sec:csmogn} and \nameref{sec:crbsmogn}.
\end{itemize}

\subsubsection{SMOTER}\label{sec:smoter}
The synthetic minority over-sampling technique (SMOTE) \citep{chawla2002smote} combines the under-sampling of frequent classes with the over-sampling of the minority class. As it is only designed for classification problems, SMOTER represents the extension of SMOTE to regression tasks \citep{torgo2013smote} and adopts large parts of it.

SMOTER divides the dataset into minority and majority datasets based on a user-defined relevance threshold, where the sample relevance values have to be bounded between 0 and 1. The number of samples in the majority subset is reduced by under-sampling with random selection of $n$ samples, where $n$ results from the ratio between majority and minority subset specified by the user. The number of samples in the minority subset is increased by over-sampling, based on SMOTE, with adjustments for continuous data.

SMOTER generates synthetic instances by interpolating between similar instances within the subset. This involves computing the $k$-nearest neighbors ($k$NN) of each minority sample, and creating new synthetics samples that lie between the minority sample and one of its nearest neighbors. The value of the target variable of the new sample is chosen based on the weighted average by the two seed samples, where the weights are determined by the distance of the new sample to the two seed samples.

As a relevance function \citet{torgo2013smote} use the interpolation between control points with pchip by \citet{ribeiro2011utility} (see Sec. \ref{sec:pchip}).

\subsubsection{SMOGN}\label{sec:smogn}
The synthetic minority over-sampling with Gaussian noise (SMOGN) method was proposed by \citet{branco2017smogn} and combines random under-sampling with either over-sampling using SMOTER or the generation of new samples based on Gaussian noise. A bounded relevance between 0 and 1 is expected. Similar to SMOTER, the dataset is partitioned into subsets containing consecutive samples in the target domain with relevance values either above or below a threshold. Partitions containing samples below the threshold are under-sampled by randomly dropping samples. Partitions containing samples above the threshold are over-sampled. The core idea for over-sampling is the same as SMOTER, but with SMOGN a new sample is only interpolated if at least one of the $k$-nearest-neighbors is "close enough" to the seed sample. If this is not the case, the new sample is created by adding Gaussian noise to the seed sample. The threshold value that is used to decide whether the neighbor is close enough is defined to be half of the median distance between the seed samples and all other samples in the partition to be over-sampled. This has the advantage that no synthetic sample is assumed in empty and probably impossible domains of the feature space between two distant seed samples. Compared to SMOTER, SMOGN can therefore be regarded as a more conservative approach. However, calculating the median distance increased the computational burden. Building multiple partitions containing consecutive samples in the target domain limits the partition size and thus eases the computational burden to calculate the distances between samples.

As a relevance function \citet{branco2017smogn} use the interpolation between control points with pchip by \citet{ribeiro2011utility} (see Sec. \ref{sec:pchip}).

\subsubsection{WERCS}\label{sec:wercs}
The weighted relevance-based combination strategy (WERCS) is a data-level imbalance mitigation strategy based on resampling, proposed by \citet{branco2019pre}.

In contrast to SMOTER and SMOGN, WERCS does not require a user defined relevance threshold to split the data, as the data are not divided into minority and majority subsets. WERCS treats all samples as candidates for either under- or over-sampling, but the probability of a sample being selected depends on its relevance (a relevance between 0 and 1 is expected). This means, that more relevant samples have higher probability of being replicated, whilst low-relevance samples have higher probability of being removed. Nevertheless, high-relevance samples can also get removed, albeit with lower probability. This approach keeps the continuous nature of regression data without forming discretized subsets. In addition, the user only has to define the amount of under- and over-sampling. 

For under- and over-sampling WERCS uses replication or dropping of the respective sample. In doing so, WERCS can be seen as one of the simplest sampling approaches, not necessitating any sort of sample imputation or interpolation.

As a relevance function \citet{branco2019pre} use the interpolation between control points with pchip by \citet{ribeiro2011utility} (see Sec. \ref{sec:pchip}).

\subsubsection{WSMOTER}\label{sec:wsmoter}
Weighting SMOTE for regression (WSMOTER) is a data-level approach that uses over-sampling to mitigate imbalance in the dataset \citep{camacho2024wsmoter}.

In contrast to SMOTER and SMOGN, WSMOTER does not split the dataset into minority and majority subsets based on a relevance threshold. Instead, similar to WERCS, each sample can be used as a seed for over-sampling. Whether a sample is selected as seed is decided by the relevance of the sample, where the relevance values have to be bounded between 0 and 1. Thus, the larger the relevance, the larger the probability of being selected for over-sampling. If a seed sample is selected, a second seed for interpolation is searched in the dataset. For this purpose, the $K$ samples are considered whose target variable values are closest to the first seed sample, where $K = 10\cdot k$. Subsequently, the $k$-nearest neighbors are determined from the $K$ samples in combined target and feature domain and one of the neighbors is randomly selected for interpolation. The actual interpolation method between the two seed samples is similar to SMOTER.

\citet{camacho2024wsmoter} use the DenseWeight method by \citet{steininger2021density} to determine the relevance of each sample (see Sec. \ref{sec:denseweight}).

\subsubsection{DenseLoss}\label{sec:denseloss}
DenseLoss is a model-agnostic, cost-sensitive learning approach introduced by \citet{steininger2021density}. 

It assumes relevance values $w_i$ between 0 and 1 and includes the relevance to weight the loss metric $m(\hat{y}_i, y_i)$ of the prediction $\hat{y}_i$ to affect the gradients of the respective training algorithm according to 
\begin{equation}
    L_\text{DenseLoss} = \frac{1}{N} \sum^N_{i=1} \phi(y_i) \cdot m(\hat{y}_i, y_i).
\end{equation}

This adaptive weighting approach ensures, that high-relevance samples contribute more prominently during model training, without artificially altering the data distribution. It effectively re-balances the training process by giving more focus to samples that would otherwise be overlooked, aiming to improve model performance in predicting values across the full target range. In not specifying the respective loss function, the approach is loss function-agnostic. In addition, it is model-agnostic, making it compatible with a variety of regression models. 

\citet{steininger2021density} propose DenseLoss together with DenseWeight as the relevance function (see Sec. \ref{sec:denseweight}).

Equivalence note: When the same relevance function $\phi(\cdot)$ is used, DenseLoss with the squared-error measure $m(\hat{y}_i, y_i) = (\hat{y}_i - y_i)^2$ is mathematically identical to SERA (see. \ref{sec:sera}).

\subsubsection{SERA}\label{sec:sera}
The Squared Error-Relevance Area (SERA) is a cost-sensitive learning approach proposed by \citet{ribeiro2020imbalanced}. It weights model predictions with respect to the relevance, penalizing errors more heavily for high-relevance targets. 

Based on a relevance function $\phi$ and a relevance cutoff $t \in [a,b]$, subsets of highly relevant cases $\mathbf{D}_t = \{\langle x_i, y_i\rangle \in \mathbf{D} \mid \phi(y_i) \ge t\}$ are defined. Here, $a$ and $b$ are the lower and upper bound of the considered relevance function. The Squared Error-Relevance for the respective cutoff is defined as 
\begin{equation}
    \text{SER}_t = \sum_{i\in\mathbf{D}^t} (\hat{y}_i - y_i)^2.
\end{equation}
By varying $t$, one obtains a monotonically decreasing SER$_t$ curve, which captures the cumulative error for increasingly relevant subsets of the data. SERA is defined as the area under the SER$_t$ curve, which is calculated by integration according to 
\begin{equation}
    \text{SERA} = \int_a^b \text{SER}_t dt  = \int_a^b \sum_{i\in\mathbf{D}^t} (\hat{y}_i - y_i)^2 dt.
\end{equation}
By integration over the complete relevance range, SERA does not require an arbitrary threshold.

As a relevance function \citet{branco2019pre} use the interpolation between control points with pchip by \citet{ribeiro2011utility} (see Sec. \ref{sec:pchip}).

\subsubsection{Probabilistic loss}\label{sec:probloss}
Probabilistic loss is a model-agnostic cost-sensitive learning approach proposed by \citet{sadouk2021novel}. 

In contrast to DenseLoss, the probabilistic loss is not calculated by scaling the loss function by the relevance $\phi(y_i)$ of the sample $i$, but by adding an additional weighted loss term to the $l_2$-loss, resulting in a regularized loss according to 
\begin{equation}
    L_\text{ProbLoss} = \frac{1}{N} \sum^N_{i=1}\left((y_i - \hat{y}_i)^2 + \sqrt{\left(\left(y_i - \hat{y}_i\right)^2 + \epsilon\right)} \cdot \phi(y_i)\right),
\end{equation}
with $\epsilon$ being a small constant ($10^{-9}$) added to ensure that the loss is differentiable at $\hat{y}_i = y_i$. The aim of this regularized loss function is to achieve a trade-off between fitting the data well and giving more importance to high-relevance cases. However, this comes with the limitation that the regularized loss only works on the basis of an $l_2$-loss. It is also assumed that the target variable is scaled between 0 and 1, as otherwise the difference between the $l_2$-loss and the regularization becomes too large \citep{sadouk2021novel}.

The probabilistic loss is proposed together with kernel density relevance as the relevance function (see Sec. \ref{sec:kernel_density_relevance}).

\subsubsection{BMC}\label{sec:bmc}
Batch-based Monte Carlo (BMC) is a cost-sensitive learning approach proposed by \citet{ren2022balanced}. It is mathematically equivalent to directly weighting the loss function, as done in DenseLoss. However, unlike other mitigation methods, \citet{ren2022balanced} assume that only the training label distribution $p_{\text{train}}(y)$ is imbalanced, while the test label distribution $p_{\text{test}}(y)$ is uniform. An interesting aspect of BMC is that it does not rely on a relevance function. Instead of explicitly weighting the loss function, BMC reformulates it as a temperature-scaled softmax function according to
\begin{equation}
    L = - \log \frac{\exp(- \|y_{\text{pred}} - y\|^2 / \tau)}{ \sum_{y' \in \text{Batch}_y} \exp(- \|y_{\text{pred}} - y'\|^2 / \tau)},
\end{equation}
where $\tau = 2\sigma^2_{\text{noise}}$ is the temperature coefficient, and $y'$ denotes the true labels in the current training batch. The noise standard deviation $\sigma_{\text{noise}}$ is treated as a hyperparameter, which can either be manually specified or learned during training. The form of loss function is equivalent to classifying within a batch.

\subsubsection{cSMOGN}\label{sec:csmogn}
The first new mitigation method we propose is continuous SMOGN (cSMOGN). It is based on the concept of SMOGN and draws together the most important features of SMOTE, SMOTER, and WSMOTER.

cSMOGN does not subdivide the dataset into minority and majority subsets, but continuously over-samples the distribution of the data in dependence on the relevance of the sample (like WSMOTER). The over-sampling is realized either by interpolation (like SMOTE) with distance-based weighting of the target variable (like SMOTER) or by replication with Gaussian noise (like SMOGN), if no second interpolation seed can be found. The space for searching for an interpolation seed is restricted by discretizing the dataset based on the method of \citet{wibbeke2022optimal}. This prevents the interpolation from creating samples over large distances in a possibly empty and unreachable space and, at the same time, limits the search space to be examined to determine the $k$-nearest neighbors' samples, which reduces the computational complexity. 

In the following the cSMOGN approach is outlined:
\begin{enumerate}
    \item Assign a value of relevance between 0 and 1 to each sample of the dataset using a respective relevance function, where a higher value indicates higher importance.
    \item Randomly select a seed sample in dependence on the squared sample relevance. Squaring the relevance gives extra emphasis to high-relevance samples as the chance to select a sample with low relevance $w_i\ll1$ gets significantly reduced. This can be illustrated by looking at a binary problem. Assuming that class 1 occurs twice as often as class 2 and therefore has half the relevance of class 2, a weighted sampling without squaring the relevance would produce the same number of samples for each of the two classes.
    \item For each over-sampling step, find the second seed sample to pair with the first seed. This is done by randomly selecting one of the $k$-nearest-neighbors of the first seed sample. However, searching the nearest neighbors over the full dataset is computationally exhaustive. Defining the subset to search for the $k$-nearest-neighbors is accomplished as follows:
    \begin{enumerate}
        \item Discretize each continuous feature of the dataset into $n_\text{bins}$ consecutive bins. Categorical features are kept discrete.
        \item Get the bin numbers of the first seed sample.
        \item Select a subset of the fully discretized dataset. The subset contains all samples which bin numbers are at maximum $\delta_b$ bins away from the seed sample bins. Categorical features have to be equal to the seed sample. This way the subset contains only samples that bare similarity to the seed sample in target as well as feature domain.
        \item  Get the undiscretized subset by using the index of the discretized subset. Subsequently, use the undiscretized subset to find the $k$-nearest neighbors.
    \end{enumerate}
    \item[4a] In case a second seed sample is found: Create the new synthetic sample by interpolation between the two seed samples using SMOTE. Calculate the target value of the synthetic sample using the weighted distance between the seed samples as in SMOTER.
    \item[4b] In case no second seed sample is found: Create the new synthetic sample by applying Gaussian noise to the seed sample as in SMOGN.
\end{enumerate}

The approach is repeated until the user-defined number of samples is reached. For better understandability we provide the pseudo-code in Alg. \ref{code:cSMOGN}.

\begin{algorithm}
\caption{Auxiliary function for cSMOGN and crbSMOGN}\label{code:auxiliary}
\begin{algorithmic}[1]
\Function{discretizeDataset}{$\mathbf{D}, n_\text{bins}$}
    \State $\mathbf{D}_\text{binned} \gets \mathbf{D}$
    \For{$d\in$ features}
        \If{\Call{IsNumeric}{$\mathbf{D}_d$}}
            \State $\mathbf{D}_{\text{binned},d} \gets$ discretize feature using $n_\text{bins}$ bins
        \EndIf
    \EndFor
    \State \Return $\mathbf{D}_\text{binned}$
\EndFunction

\Statex
\Function{gaussianNoise}{$\mathbf{D}, s_1, \delta_n$}
    \For{$d\in$ features}
        \If{\Call{IsNumeric}{$\mathbf{D}_d$}}
            \State $s_{new, d} \gets s_{1,d} + N\left(0, \delta_n \cdot \text{std}(\mathbf{D}_d)\right)$ // std: standard deviation
        \Else
            \State $s_{new, d} \gets s_{1,d}$
        \EndIf
    \EndFor
    \State \Return $s_\text{new}$
\EndFunction

\Statex
\Function{interpolate}{$s_1, s_2$}
    \For{$d\in \text{features} \setminus \{\textit{target}\}$}
        \If{\Call{IsNumeric}{$s_{1,d}$}}
            \State $s_{\text{new},d} \gets s_{1,d} + \Call{RANDOM}{0,1} \cdot \left(s_{1,d} - s_{2,d}\right)$
        \Else
            \State $s_{\text{new},d} \gets$ randomly select among $s_{1,d}$ and $s_{2,d}$
        \EndIf
    \EndFor
    \State $\textit{dist}_1 \gets \Call{Distance}{s_1, s_\text{new}}$
    \State $\textit{dist}_2 \gets \Call{Distance}{s_2, s_\text{new}}$
    \State $s_{\text{new}, target} = \frac{\textit{dist}_1 \cdot s_{2,target} + \textit{dist}_2 \cdot s_{s,target}}{\textit{dist}_1 + \textit{dist}_2}$
    \State \Return $s_\text{new}$
\EndFunction

\Statex
\Function{similarSamples}{$\mathbf{D}, \mathbf{D}_\text{binned}, s_1, \delta_b$}
    \State $s \gets$ get discretized seed sample $\mathbf{D}_\text{binned}[s_1]$
    \State $e \gets \{\}$
    \For{$i\in$ samples}
        \For{$d\in$ features}
            \If{\textbf{not} $\mathbf{D}_\text{binned}[i,d] \geq (s[d] - \delta_b)$ \textbf{and} $\mathbf{D}_\text{binned}[i,d] \leq (s[d] + \delta_b)$}
                \State $e \gets e\cup i$
                \State \textbf{break}
            \EndIf
        \EndFor
    \EndFor
    \State $s_\text{similar} \gets \mathbf{D}[\sim e]$
    \State \Return $s_\text{similar}$
\EndFunction
\end{algorithmic}
\end{algorithm}

\begin{algorithm}
\caption{Pseudocode for cSMOGN}\label{code:cSMOGN}
\begin{algorithmic}[1]
\Function{cSMOGN}{$\mathbf{D}, \phi, n_\text{bins}, \delta_b, n_\text{sample}, \delta_n, k$}
    \State \textbf{Input:}
    \State \quad $\mathbf{D}$ - Dataset
    \State \quad $\phi$ - Relevance values of the samples in $\mathbf{D}$ with $\phi \in [0,1]$
    \State \quad $n_\text{bins}$ - Number of bins to discretize $\mathbf{D}$ in
    \State \quad $\delta_b$ - Allowed bin distance
    \State \quad $n_\text{sample}$ - Number of samples to generate
    \State \quad $\delta_n$ - Noise level
    \State \quad $k$ - Number of neighbors
    \State \textbf{Output:} $\mathbf{D}_{\text{resampled}}$ 
    % Start algorithms
    \State $\mathbf{D}_{\text{resampled}} \gets \mathbf{D}$
    \State $\mathbf{D}_\text{binned} \gets$\Call{discretizeDataset}{$\mathbf{D}, n_\text{bins}$}
    \Comment{See Algorithm \ref{code:auxiliary}}
    \State $i \gets 0$
    \While{$i < n_\text{sample}$}
        \State $s_1, w_1 \gets$ randomly select one sample from $\mathbf{D}$ and the respective relevance from $w$
        \If{$w_1^2 >$ random uniform number between 0 and 1}
            \State $s_\text{similar} \gets$\Call{similarSamples}{$\mathbf{D}, \mathbf{D}_\text{binned}, s_1, \delta_b$}
            \Comment{See Alg. \ref{code:auxiliary}}
            \If{$s_\text{similar} = None$}
                \State $s_\text{new} \gets$ \Call{gaussianNoise}{$\mathbf{D}, s_1, \delta_n$}
                \Comment{See Alg. \ref{code:auxiliary}}
            \Else
                \State $nns \gets$ $\text{kNN}(s_1, s_\text{similar}, k)$ // $k$-Nearest Neighbors of $s_1$ in $s_\text{similar}$
                \State $s_2 \gets$ randomly choose one of the $nns$
                \State $s_\text{new} \gets$ \Call{interpolate}{$s_1, s_2$}
                \Comment{See Alg. \ref{code:auxiliary}}
            \EndIf
            \State $\mathbf{D}_{\text{resampled}} \gets \mathbf{D}_{\text{resampled}} \bigcup s_\text{new}$
            \State $i++$
        \EndIf
    \EndWhile   
    \State \Return $\mathbf{D}_{\text{\text{resampled}}}$
\EndFunction
\end{algorithmic}
\end{algorithm}

To be able to run cSMOGN for continuous and categorical variables, we propose to use the heterogeneous Euclidean-overlap metric (HEOM) by \citet{wilson1997improved} for all distance calculations.

Theoretically, cSMOGN can be used with any relevance function scaled between 0 and 1, however, we propose using the density-distance relevance function (see Sec. \ref{sec:density_dist_relevance}).

\subsubsection{crbSMOGN}\label{sec:crbsmogn}
The second new mitigation method we propose is the sampling technique continuous ratio-based SMOGN (crbSMOGN). In contrast to all other mitigation methods, crbSMOGN does not require sample relevance values scaled between 0 and 1 but relies on a ratio-based relevance. Ratio-based relevance values express the importance in relativity to the nominal relevance of $w_i = 1$, thus, for example, a sample with relevance $w_i = 2$ is twice as relevant.

In principle, the over-sampling in crbSMOGN works similarly to cSMOGN. However, due to the high expressiveness of the ratio-based relevance function, crbSMOGN does not require a user-defined input to specify the number of samples to create. crbSMOGN obtains the number of samples to be over-sampled from the ratio-based relevance values. The data are sampled until the entire target domain has an equal relevance of 1. However, whether this is possible depends on the distribution of the data, as it is not possible for crbSMOGN to generate data in areas where none exist.

The over-sampling is realized by either interpolation (like SMOTE) with distance-based weighting of the target variable (like SMOTER) or by replication with Gaussian noise (like SMOGN), if no second interpolation seed can be found. The space for searching for an interpolation seed is restricted by discretizing the dataset, based on the method of \citet{wibbeke2022optimal}. This prevents the interpolation from creating samples over large distances in a possibly empty and unreachable space and, at the same time, limits the search space to be examined to determine the $k$-nearest neighbors' samples, which reduces the computational complexity.

In the following, the crbSMOGN approach is outlined:
\begin{enumerate}
    \item Assign a value of relevance to each sample of the dataset using a respective ratio-based relevance function.
    \item Over-sample each sample with a ratio-based relevance $w_i> 1$ in accordance to its relevance. This means that a sample with $w_i = 1.5$ has 50\% chance of being over-sampled and a sample with $w_i = 3$ is over-sampled twice.
    \item The over-sampling is done with interpolation. Thus, a second seed sample to pair with the first seed is needed. This is done by selecting the $k$-nearest-neighbors of the first seed sample, where $k$ is the over-sampling rate of first seed. However, searching the nearest neighbors over the full dataset is computationally exhaustive. Defining the subset to search for the $k$-nearest-neighbors is accomplished as follows (similar to cSMOGN):
    \begin{enumerate}
        \item Discretize each continuous feature of the dataset into $n_\text{bins}$ bins. Categorical features are kept discrete.
        \item Get the bin numbers of the first seed sample.
        \item Select a subset of the full discretized dataset. The subset contains all samples which bin numbers are at maximum $\delta_b$ bins away from the seed sample bins. Categorical features have to be equal to the seed sample. This way the subset contains only samples that bare similarity to the seed sample in target as well as feature domain.
        \item Get the undiscretized subset by using the index of the discretized subset.
        \item If the subset is larger than the over-sampling rate of the seed sample return the $k$-nearest neighbors in the subset with regard to the seed sample, where $k$ is the over-sampling rate. If the subset is smaller than the over-sampling rate of the seed sample return the full subset. In this case, the calculation of the nearest neighbors is omitted, as the discretization already ensures that all samples in the subset are similar to the seed.
    \end{enumerate}
    \item[4a] In case the subset is empty: Create the new synthetic samples by applying Gaussian noise to the seed sample as in SMOGN.
    \item[4b] In case the subset contains $k$ samples: Create a new synthetic samples for each sample in the subset by interpolation between the seed sample using SMOTE. Calculate the target values of the synthetic samples using the weighted distance between the seed sample and the neighbors as in SMOTER.
    \item[4b] In case the subset contains less than $k$ samples: Create a new synthetic sample by interpolation for each sample in the subset. In addition, create synthetic samples by applying Gaussian noise to the seed sample until in total $k$ samples are created.
\end{enumerate}

For better understandability we provide the pseudocode in Alg. \ref{code:crbSMOGN}.

\begin{algorithm}
\caption{Pseudocode for crbSMOGN}\label{code:crbSMOGN}
\begin{algorithmic}[1]
\Function{crbSMOGN}{$\mathbf{D},  w, n_\text{bins}, \delta_b, \delta_n$}
    \State \textbf{Input:}
    \State \quad $\mathbf{D}$ - Dataset
    \State \quad $w$ - Relevance values of the samples in $\mathbf{D}$
    \State \quad $n_\text{bins}$ - Number of bins to discretize $\mathbf{D}$ in
    \State \quad $\delta_b$ - Allowed bin distance
    \State \quad $\delta_n$ - Noise level
    \State \textbf{Output:} $\mathbf{D}_{\text{resampled}}$ 
    % Start algorithms
    \State $\mathbf{D}_{\text{resampled}} \gets \mathbf{D}$
    \State $\mathbf{D}_\text{binned} \gets$\Call{discretizeDataset}{$\mathbf{D}, n_\text{bins}$}
    \Comment{See Alg. \ref{code:auxiliary}}
    \For{$i \in$ samples in $\mathbf{D}$}
        \State $s_1, w_1 \gets (y_i,\mathbf{x}_i), w_i$ \Comment{sample $i$ and the respective relevance}
        \If{$(w_1 \% 1) > $ \Call{random}{0,1}}
            \State $r = $ \Call{Ceil}{$w_i$} - 1
        \Else
            \State $r = $ \Call{Floor}{$w_i$} - 1
        \EndIf
        \If{$r > 0$}
        \State $s_\text{similar} \gets$\Call{similarSamples}{$\mathbf{D}, \mathbf{D}_\text{binned}, s_1, \delta_b$}
        \Comment{See Alg. \ref{code:auxiliary}}
        \If{\Call{size}{$s_\text{similar}$} $ \geq r$}
            \State $n_\text{gauss} \gets 0$
            \State $nns \gets$ \Call{kNN}{$s_1, s_\text{similar}, k=r$} \Comment{$k$-Nearest Neighbors of $s_1$ in $s_\text{similar}$}
        \Else
            \State $n_\text{gauss} \gets r -$ \Call{size}{$s_\text{similar}$}
            \State $nns \gets s_\text{similar}$            
        \EndIf
            \For{$s \in nns$}
                \State $s_\text{new} \gets$ \Call{interpolate}{$s_1, s_2$}
                \Comment{See Alg. \ref{code:auxiliary}}
                \State $\mathbf{D}_{\text{resampled}} \gets \mathbf{D}_{\text{resampled}} \bigcup s_\text{new}$
            \EndFor
            \For{$n=1$ to $n_\text{gauss}$}
                \State $s_\text{new} \gets$ \Call{gaussianNoise}{$\mathbf{D}, s_1, \delta_n$}
                \Comment{See Alg. \ref{code:auxiliary}}
                \State $\mathbf{D}_{\text{resampled}} \gets \mathbf{D}_{\text{resampled}} \bigcup s_\text{new}$
            \EndFor
        \EndIf
    \EndFor   
    \State \Return $\mathbf{D}_{\text{resampled}}$
\EndFunction
\end{algorithmic}
\end{algorithm}

crbSMOGN can be used with any ratio-based relevance function, however, to the best of our knowledge the newly proposed density-ratio relevance (see Sec. \ref{sec:density_ratio_relevance}) is the only known one, so far.

\subsection{Evaluation metrics for imbalanced regression}\label{sec:evaluation_metrics}
Standard regression metrics, such as Mean Squared Error (MSE) or Mean Absolute Error (MAE), assume that all prediction errors are equally important across the domain of the target variable. In imbalanced regression tasks, this assumption is often violated: extreme values are typically under-represented and may be of higher relevance, while frequent values dominate the overall error. As a consequence, traditional metrics fail to discriminate between models that perform well on rare, high-impact cases and those that perform well only on the abundant, low-impact cases. 

Unlike classification, where metrics such as precision, recall, and the $F_1$-score naturally account for class imbalance, regression lacks the discrete notion of classes. This makes the direct adaptation of classification metrics non-trivial. In the following, multiple types of metrics for the evaluation of imbalanced regression tasks are presented, including:
\begin{itemize}
    \item \nameref{sec:diagram_based_eval}, \nameref{sec:precision_and_recall}, \nameref{sec:loss_from_cost_sensitive} and \nameref{sec:partitioned_loss_metrics}.
\end{itemize}

\subsubsection{Diagram-based evaluation} \label{sec:diagram_based_eval}
The Receiver Operating Characteristic (ROC) analysis is a well-established method for graphically evaluating model performance in classification tasks. It plots the false-positive rate on the x-axis and the true-positive rate on the y-axis. In regression, curve-based metrics are used to visualize performance across a continuum of operating points using diagrams. However, a direct adaptation of ROC analysis to regression does not exist. Instead, quantities such as error tolerances, cost ratios, or relevance thresholds are used to derive ROC-like diagrams. By inspecting the curve shape, these measures enable the assessment of how well a model performs in the minority region of the respective distribution.

Notable examples of diagram-based evaluation metrics include:
\begin{itemize}
    \item Regression Error Characteristic (REC): plots the cumulative fraction of predictions within an increasing error tolerance threshold, as introduced by \citet{bi_regression_2003}.
    \item Regression Error Characteristic Surfaces (RECS): extends REC diagrams by incorporating the target value as a third dimension, highlighting model performance across the target domain, as proposed by \citet{torgo_regression_2005}.
    \item Regression ROC (RROC): plots the total over-estimation on the x-axis and the total underestimation on the y-axis, as described by \citet{hernandez-orallo_roc_2013}.
\end{itemize}

Furthermore, many other metrics, including those used in cost-sensitive learning, can be represented as a diagrams plotted over a variable relevance threshold. This methodology has been applied, for example by \cite{ribeiro2020imbalanced} using the SERA metric.

While visualization can be a powerful tool for understanding model performance, diagram-based metrics prioritize visualization over quantitative evaluation, which can be a limitation.

\subsubsection{Precision and recall for regression} \label{sec:precision_and_recall}
\cite{torgo2009precision} extend the concepts of precision and recall from classification to regression. This approach involves dividing the dataset into minority and majority subsets based on a relevance threshold. Predictions that fall within a specified error tolerance of the true value are considered correct, creating a binary classification. This enables the computation of precision and recall in a regression context, which can be combined into an $F_1$-measure.

However, this method has a significant limitation: it requires the dataset to be artificially split into two subsets based on a user-defined relevance threshold to create a minority and a majority subset. This can be a disadvantage, as the choice of threshold can be subjective and may not reflect the underlying characteristics of the data. In addition, it neglects the continuous nature of regression data.

\subsubsection{Loss functions from cost-sensitive learning} \label{sec:loss_from_cost_sensitive}
In cost-sensitive learning, the distribution of data is not changed, but rather the training is modified. To do this, the loss function is adjusted or scaled. To assess the performance of a model the loss functions can also be used as evaluation metrics.

DenseLoss \citep{steininger2021density} is a scalar metric that can be employed directly as a loss function. It estimates the expected prediction error over the target domain while weighting each sample according to a relevance function. A full presentation of DenseLoss is provided in Section~\ref{sec:denseloss}.

The Squared Error‑Relevance Area (SERA) \citep{ribeiro2020imbalanced} generalises the ordinary squared‑error loss by applying relevance weighting to the individual errors. By evaluating the loss for a series of relevance cut‑off thresholds, SERA yields a monotonic SER$_t$ curve that can be used for diagram‑based evaluation. When the loss is integrated over the entire relevance range, the resulting SERA value is mathematically equivalent to the weighted‑squared‑error formulation of DenseLoss. A complete description of SERA is provided in Section~\ref{sec:sera}.

Probabilistic loss \citep{sadouk2021novel} is a model-agnostic cost-sensitive learning approach that modifies the standard loss by adding a relevance-weighted term rather than simply scaling the loss by relevance. This regularization emphasizes high-relevance samples while still fitting the overall data, creating a trade-off between global accuracy and targeted performance. A complete discussion is provided in Section~\ref{sec:probloss}.

Batch-based Monte Carlo (BMC) \citep{ren2022balanced} is a cost-sensitive learning approach designed for scenarios where the training label distribution is imbalanced but the test distribution is uniform. Unlike other methods, BMC does not rely on an explicit relevance function. Instead, it reformulates the loss to focus on predictions relative to other samples within each training batch, effectively emphasizing underrepresented targets. A full description is provided in Section~\ref{sec:bmc}.

\subsubsection{Partitioned loss metrics} \label{sec:partitioned_loss_metrics}
Standard regression loss functions, such as MSE, MAE, or RMSE, can be adapted for imbalanced regression by evaluating them on subsets of the target variable. The target range is partitioned into bins, and the loss is computed separately for each bin. In general, bins may be defined according to any user-specified rule. Frequency-based partitions are common in the literature \citep{steininger2021density, camacho2024wsmoter}, but relevance-driven or domain-driven partitions may be more appropriate when the practical importance of certain target regions is not aligned with their empirical density.

The bin-based evaluation yields a set of metric values, each reflecting model performance on a specific segment of the target domain. The per-bin losses can be further aggregated (e.g., by computing their unweighted mean) to equalize the contribution of each bin to the overall metric, thereby counteracting the dominance of high-density regions.

A critical aspect concerns the choice of bin boundaries. Small shifts in the partitioning scheme may move individual observations between adjacent bins, potentially leading to large changes in the estimated error for sparsely populated partitions. This sensitivity is amplified when bins contain only a small number of observations. Consequently, there exists an inherent trade-off between granularity and statistical robustness: partitions should be sufficiently narrow to capture local performance variations, yet sufficiently broad to ensure a meaningful number of samples per bin.

Partitioned loss metrics do not inherently require an explicit relevance function. When partitions are constructed using quartiles or fixed intervals of the target domain (as done by \citet{steininger2021density, camacho2024wsmoter}) the procedure relies solely on the empirical distribution of the target variable and is therefore independent of any externally specified importance weighting. In this form, partitioned metrics provide an intuitive and flexible evaluation scheme that enables direct comparison of mitigation strategies without committing to a particular relevance formulation.

However, if the partitioning criterion itself is derived from a relevance function, then the resulting metric implicitly reflects this relevance definition. In such cases, the evaluation is no longer independent of the chosen notion of relevance. Thus, while partitioned loss metrics can be relevance-free, their dependence on a relevance function ultimately depends on how the partitions are defined.

\subsection{Ensembles}\label{sec:ensemble_formation}
In many applications, it has been found that forming an ensemble from several models leads to better results \citep{galar2011review, moniz2017evaluation, moniz2018smoteboost, branco2018rebagg}. The aim of an ensemble is to balance out the disadvantages of individual models, for example by forming a weighted average of all models.

We propose to form an ensemble of a "normal" model $M_\text{normal}$ trained without any imbalance mitigation strategy and a model $M_\text{imbalanced}$ that has been adjusted for data imbalance. To obtain the final approximation $\hat{y}_i$ of the ensemble, we propose to use either the mean of both models, the weighted mean or select one model based on a relevance threshold.

The mean of both models can be formed according to 
\begin{equation} \label{eqn:ensemble_mean}
    \hat{y}_{\text{mean}, i} = \frac{M_\text{imbalanced}(\mathbf{x}_i) + M_\text{normal}(\mathbf{x}_i)}{2}.
\end{equation}

The weighted mean of an ensemble of the two models can be formed according to

\begin{equation} \label{eqn:ensemble_ratio}
    \hat{y}_{\text{weighted}, i} =\frac{\beta_1(y_i) \cdot M_\text{imbalanced}(\mathbf{x}_i) + \beta_2(y_i)\cdot M_\text{normal}(\mathbf{x}_i)}{\beta_1(y_i) + \beta_2(y_i)},
\end{equation}
where $\beta_1(y_i)$ and $\beta_2(y_i)$ are the positive weights of the models considering the respective sample $y_i$, which can be substituted by the relevance value.

Under the assumption that $\beta_2$ being the complement of $\beta_1$, leading to $\beta_1(y_i) + \beta_2(y_i) = 1$ and $\beta_1(y_i) = w_\text{scaled}(y_i)$, Ep. \ref{eqn:ensemble_ratio} can be simplified to
\begin{equation}
    \hat{y}_{\text{weighted}, i} = w_{\text{scaled},i} \cdot M_\text{imbalanced}(\mathbf{x}_i) + \left(1-w_{\text{scaled},i}\right) \cdot M_\text{normal}(\mathbf{x}_i),
\end{equation} 
where $w_{\text{scaled},i}$ denotes positively scaled relevance values bounded between 0 and 1.

Now the question arises of how to obtain $y_i$ if the true value of $y_i$ is unknown during inference. On the one hand, $y_i = M_\text{imbalanced}(\mathbf{x}_i)$ would result in a bias towards $M_\text{imbalanced}$, on the other hand $y_i = M_\text{normal}(\mathbf{x}_i)$ biases towards $M_\text{normal}$. We therefore propose to use the mean according to Eq. \ref{eqn:ensemble_mean}, to mitigate the issue. Resulting in
\begin{equation}
    \hat{y}_{\text{weighted}, i} = w_\text{scaled}(\hat{y}_{\text{mean}, i}) \cdot M_\text{imbalanced}(\mathbf{x}_i) + \left(1-w_\text{scaled}\left(\hat{y}_{\text{mean}, i}\right)\right)\cdot M_\text{normal}(\mathbf{x}_i).
\end{equation}

Considering ratio-based relevance, the ratio between the two models is already included in the relevance values $w_\text{ratio}(y_i)$. Thus, under the assumption that $\beta_1(y_i) = w_\text{ratio}(\hat{y}_{\text{mean}, i})$ and $\beta_2(y_i) = 1$ the ensemble equation \ref{eqn:ensemble_ratio} can be simplified according to
\begin{equation}
    \hat{y}_{\text{weighted}, i} = \frac{w_\text{ratio}(\hat{y}_{\text{mean}, i})\cdot M_\text{imbalanced}(\mathbf{x}_i) + M_\text{normal}(\mathbf{x}_i)}{w_\text{ratio}(\hat{y}_{\text{mean}, i})+1}.
\end{equation}

Considering the ensemble based on a relevance threshold, density-distance relevance and density-ratio relevance offer improved interpretability. Both provide an equilibrium point at which the targets' empirical density is equal to the domain-priority density. This point ($w_\text{dist} = 0.5$, $w_\text{ratio} = 1$) can be used as a threshold $t$ for model selection in an ensemble according to 
\begin{equation} \label{eqn:model_weight}
\hat{y}_{\text{threshold}, i} = 
\begin{cases} 
M_\text{normal}(\mathbf{x}_i) & \text{if } w(\hat{y}_{\text{mean}, i}) < t \\
M_\text{imbalanced}(\mathbf{x}_i) & \text{if } w(\hat{y}_{\text{mean}, i}) \geq t,
\end{cases}
\end{equation}
where $t$ is the set relevance threshold. For other relevance functions scaled between 0 and 1 we suggest to use $t=0.5$.

\section{Experimental analysis} \label{sec:experiment}
In this section we evaluate how different combinations of relevance functions and mitigation methods affect sampling and the model bias caused by data imbalance.

First, we demonstrate the impact of sampling-based mitigation strategies on the target variable in a qualitative analysis. The two subsections cover:
\begin{itemize}
    \item \nameref{sec:dual-source-relevance-functions}\\
        It is shown that by using the empirical distribution and the domain-priority distribution within the newly proposed dual-source relevance functions (density-distance and density-ratio relevance), the distribution of the target variable can be specifically adapted to the respective use case.
    \item \nameref{sec:comparison-of-mitigation-methods}\\
        It is shown that different mitigation methods have different effects on the distribution of the target variable, where only crbSMOGN can achieve close-to-uniform distribution.
\end{itemize}

Following the qualitative analysis, we evaluate the performance of various strategies to avoid model bias in a quantitative benchmark study. For this purpose, combinations of relevance functions and mitigation methods are tested and the performance of the individual strategies is compared. More specifically, the sampling methods SMOGN, WERCS, WSMOTER, cSMOGN and crbSMOGN as well as the cost-sensitive learning methods DenseLoss and probabilistic loss are examined. In addition, all relevance functions presented in Sec. \ref{sec:relevance_function} are analyzed. The quantitative benchmark is split into three parts:
\begin{itemize}
    \item \nameref{sec:case_study_synthetic}\\
        In a benchmark with synthetic datasets, it is shown that all efforts to increase performance for rare samples lead to a decrease in performance for frequent samples. In addition, it is shown that resampling with crbSMOGN yields the best performance for rare samples.
    \item \nameref{sec:case_study_real_world_data}\\
        In a benchmark with real-world datasets, it is shown, that resampling with the newly proposed cSMOGN and crbSMOGN achieves good performance for rare samples of real-world datasets. However, the performance of mitigation strategies in general is model type dependent. In addition, it is shown that the trade-off between performance gains for rare samples and performance losses for frequent samples persists, regardless of the mitigation strategy.
    \item \nameref{sec:case_study_ensembles}\\
        In a benchmark of the proposed ensemble methods, it is shown that the mean-based ensembles reduce the overall impact of the performance trade-off. They reduce the performance decrease for frequent samples while preserving reasonable performance for rare samples.
\end{itemize}

To estimate the empirical distribution, most relevance functions use kernel density estimation, with Silverman's rule of thumb for bandwidth estimation. Silverman's rule of thumb is a non-optimal estimator for non-normal distributions as it tends to over-smooth non-normally or even non-parametric distributed target value domains. A more precise solution is the improved Sheather-Jones (ISJ) algorithm \citep{botev2010kernel}. However, up to now, all density-based relevance functions used the Silverman bandwidth, due to its simplicity. Thus, we refrained from using the ISJ bandwidth as it is much more susceptible to domain scaling than Silverman's rule of thumb and does not converge to a solution for any given empirical distribution. 

For the relevance by interpolation with control points the Python implementation from \citet{kunz2020smogn} is used\footnote{\url{https://github.com/nickkunz/smogn}}. For DenseWeight the implementation of the authors is used as provided in the article \citep{steininger2021density}. The code for the newly proposed relevance functions density-distance relevance and density-ratio relevance in addition to cSMOGN and crbSMOGN are provided as a Python package on GitHub\footnote{\url{https://github.com/OFFIS-ROC/imbami}}. The code to rerun the complete experimental analysis is provided on GitHub\footnote{\url{github url to be added.}}

\subsection[Impact of sampling on the target variable]{Impact of sampling-based mitigation strategies on the target variable} \label{sec:impact-of-sampling}
To analyze the impact of sampling-based mitigation strategies on the target variable we investigate the change of the distribution of the target variable after resampling with different mitigation strategies. First, we investigate the behavior of the newly introduced dual‑source relevance functions density‑distance and density‑ratio relevance using both source distributions. In the second part we compare the sampling behavior to other single source mitigation strategies.

\subsubsection{Dual-source relevance functions} \label{sec:dual-source-relevance-functions}
Dual-source relevance functions combine information from the empirical distribution of the target variable with a user‑defined domain-priority distribution. By design they reduce to single‑source relevance functions when the complementary distribution is set to uniform, i.e., when only empirical frequency (or only domain priority) considered. Consequently, dual‑source relevance eliminates the need for relevance thresholds that are required in many single‑source approaches to distinguish between over- or under-represented regions.

\begin{figure}
    \centering
    \includegraphics[width=1\textwidth]{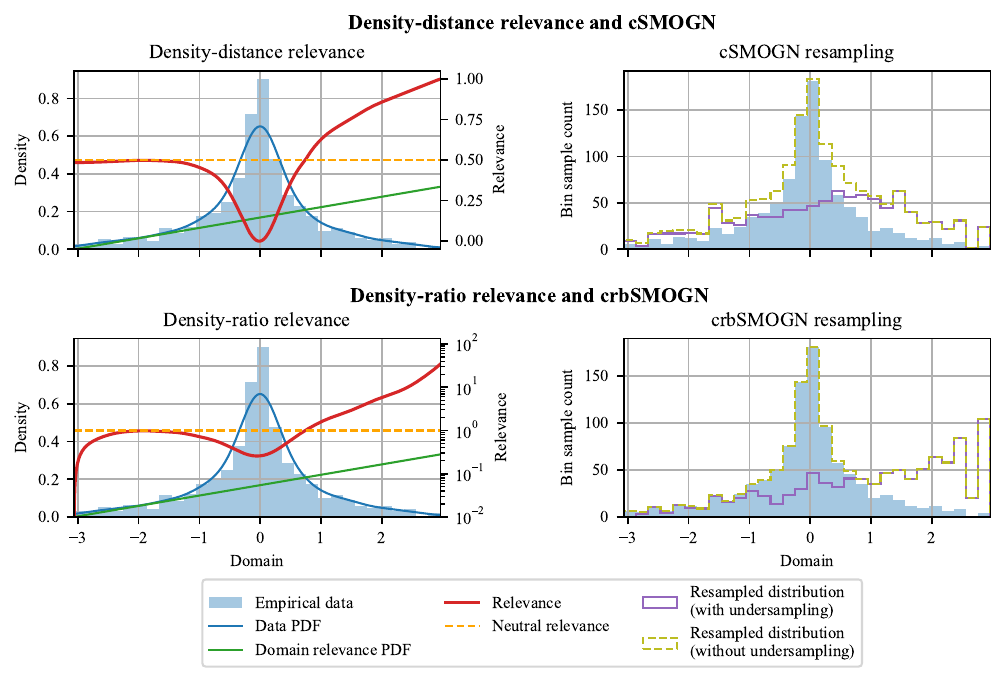}
    \caption{Behavior of domain- and empirical frequency-guided relevance functions for imbalanced regression on a Nernst equation dataset. (Left) Empirical data distribution (blue) and linearly rising domain priority (green) are fused into relevance functions (red), where neutral relevance (orange) defines the decision boundary: regions requiring no resampling when domain and empirical distributions align. (Right) Resampling behavior confirms the mechanism: relevance > neutral drives over-sampling (high-relevance regions), while relevance < neutral triggers under-sampling (low-relevance regions), aligning synthetic samples with domain priorities in Nernst equation data where higher electrochemical potentials are prioritized by the rising domain priority.}
    \label{fig:combined_relevance}
\end{figure}

Figure \ref{fig:combined_relevance} illustrates this concept using the Nernst equation dataset. The left panel shows how normally distributed empirical data (blue) and linearly rising domain priority (green) combine to form relevance functions (red). The neutral relevance threshold (orange) defines the boundary between regions requiring over-sampling (relevance > neutral) and under-sampling (relevance < neutral). The right panel demonstrates the resampling behavior using cSMOGN and crbSMOGN. Here, high-relevance regions are over-sampled and low-relevance regions are under-sampled to align with domain priorities.

This dual-source approach offers two key advantages: (1) It generalizes single-source functions when one distribution is uniform, and (2) it eliminates the need for artificial relevance thresholds by naturally distinguishing between over-sampling and under-sampling regions based on the interplay between empirical frequency and domain priority.

\subsubsection{Comparison of mitigation methods}\label{sec:comparison-of-mitigation-methods}
\begin{figure}
    \centering
    \includegraphics[width=1\textwidth]{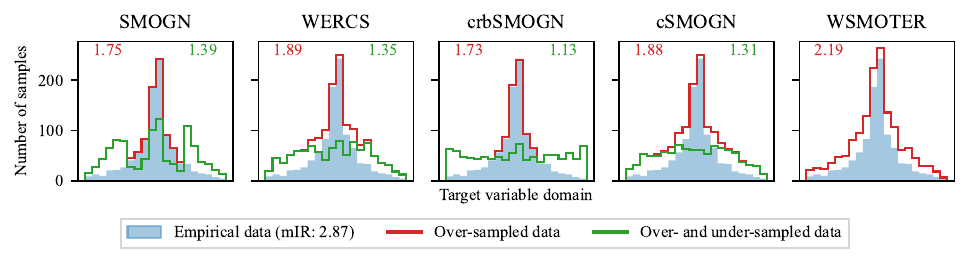}
    \caption{Comparison of the effect of different data imbalance mitigation methods on the target variable of the synthetic Nernst dataset. Shown are the original distribution of the data, the distribution after over-sampling and the distribution after over- and under-sampling. The colored values indicate the mean imbalance ratio (mIR) of the over-sampled data (red) and the over- and under-sampled data (green).}
    \label{fig:influence_on_target}
\end{figure}

To highlight the differences between the individual data imbalance mitigation methods, the effect on the target variable of the dataset is examined. The target variable is decisive for loss functions during model training, which makes it particularly important. For this purpose, five mitigation methods are applied to the synthetic Nernst dataset (see Appendix \ref{sec:appendix_data} Table \ref{tab:synthetic_data}). As relevance functions, the functions used in the respective publication of the methods are used. These are SMOGN and WERCS with control points interpolation using pchip, WSMOTER with DenseWeight, cSMOGN with distance-based relevance and crbSMOGN with ratio-based relevance. The result is shown in Fig. \ref{fig:influence_on_target}. No under-sampling could be carried out for WSMOTER, as this is not intended in the method.

The figure shows that, as intended, under-sampling essentially affects the areas of very frequently occurring samples. SMOGN is the only method that divides the dataset into a majority and minority subset. Since all samples in the minority subset are over-sampled to the same extent, data points that lie on the boundary between the subsets are represented significantly more frequently after sampling. This leads to the two dips that can be seen in the distribution.

Furthermore, it is notable that WSMOTER exhibits a relatively consistent rate of introducing new samples to the domain, irrespective of the frequency of similar existing samples. This behavior stems from the fact that the probability of generating a new sample in WSMOTER is linearly proportional to the sample's relevance, which, due to the implementation of DenseWeight, is directly tied to the sample's frequency. Consequently, while a frequent sample might be over-sampled at half the rate of a less frequent sample, its higher prevalence in the dataset causes the absolute number of over-sampled instances to be similar. This effect is circumvented by cSMOGN, which employs a sampling probability dependent on the square root of the sample relevance. This modification results in a larger concentration of generated samples within the region of high-relevance samples.

All of the methods presented use the empirical distribution of the data as the basis for relevance estimation, but only crbSMOGN succeeds in obtaining a (virtually) uniform distribution through sampling. However, this is only possible because the domain is completely covered by the data record. If there are gaps in the domain, these cannot be filled by crbSMOGN either.

The mean imbalance ratio (mIR) provides a value to assess the average imbalance in the data. A mIR = 1 indicates that the empirical distribution matches the relevance distribution (in this case uniform) and the dataset is not imbalanced. The imbalance increases with increasing mIR, where, for example, mIR = 2 denotes that the frequencies of the samples deviate on average by 2-times from the expected uniform frequency \citep{wibbeke2024quantification}. Since under-sampled datasets are closer to a uniform distribution, the mIR values are generally lower than those for datasets without under-sampling. However, all presented approaches decrease the mIR. Based on the calculated mIR values, crbSMOGN with under-sampling should be the best method to avoid data imbalance.

\subsection{Benchmark of mitigation strategies}\label{sec:performance_evaluation}
In this section the performance of various relevance function and mitigation method combinations is evaluated on synthetic and real-world datasets. We speak of synthetic data if it is created artificially. The functional relationship between the target variable and the input features as well as noise levels are known. On the other hand, real-world data has unknown noise levels and is potentially not based on parametric/regular distributions, which makes modeling more difficult.

The list of the real-world datasets used is provided in Appendix \ref{sec:appendix_data} Tab. \ref{tab:datasets}.  The datasets cover a wide range of applications. They range between several hundred to thousands of samples and between 2 and 149 features. The list containing all synthetic datasets is provided in Appendix \ref{sec:appendix_data} Tab.~\ref{tab:synthetic_data}. 

All datasets are randomly split into training (70\%) and testing (30\%) sets. When using imbalanced data, it is important that the training and testing data have comparable distributions. Therefore, each split is performed 100 times and the similarity of the target value distribution of the subsets is evaluated using the mean imbalance ratio (mIR) \citep{wibbeke2024quantification}\footnote{\url{https://github.com/OFFIS-ROC/imbaqu}}. Only the most similar (lowest mIR) split is kept and used for model training.

All combinations of relevance functions and mitigation methods shown in Table \ref{tab:combination} are tested. In addition, all sampling-based mitigation methods (SMOGN, WERCS, cSMOGN and crbSMOGN) are evaluated as pure over-samplers and as combined over- and under-samplers. This results in a total of 46 different combinations, plus one model per model type without using any imbalance mitigation strategy for comparability. This will further on be referred to as the no-mitigation baseline. As models, multilayer perceptron (MLP) feed forward neural networks, extreme gradient boosting trees (XGBoost) and Random Forests (Forest) are used. This results in 141 different trained models per dataset for one experiment run. To make a statistical statement, the training was repeated 50 times.

% 141 = (46+1)*3

The various hyper-parameters for the models, relevance functions and mitigation methods can be found in Appendix~\ref{sec:appendix_hyperparameter}.

\begin{table}[ht]
\centering
\renewcommand{\arraystretch}{1.2}
\begin{tabularx}{\textwidth}{
    p{0.3cm}
    |m{2cm}
    |>{\centering\arraybackslash}m{1.4cm}
    |>{\centering\arraybackslash}m{1.9cm}
    |>{\centering\arraybackslash}m{1.4cm}
    |>{\centering\arraybackslash}m{1.4cm}
    |>{\centering\arraybackslash}m{1.4cm}
    |>{\centering\arraybackslash}m{1.4cm}
    |>{\centering\arraybackslash}m{1.4cm}|}
\multicolumn{2}{c}{}&\multicolumn{7}{|c|}{Relevance function} \\
\cline{3-9}
\multicolumn{2}{c|}{}&Inter-polation &Histogram-based&Dist. smoothing& Kernel density relevance & Dense\-Weight &Density-distance relevance &Density-ratio relevance\\
\cline{1-9}
\multirow{7}{*}[-0.3cm]{\rotatebox{90}{ Mitigation method}}
&SMOGN&\checkmark&\checkmark&\checkmark&\checkmark& *1 & \checkmark & - \\
\cline{2-9}
&WERCS&\checkmark&\checkmark&\checkmark&\checkmark& *1 & \checkmark & - \\
\cline{2-9}
&WSMOTER&*2&&&*1& \checkmark& *1 & -  \\
\cline{2-9}
&DenseLoss&\checkmark&\checkmark&\checkmark&\checkmark& \checkmark & \checkmark & \checkmark \\
\cline{2-9}
%&SERA & \checkmark&\checkmark&\checkmark&\checkmark& \checkmark & \checkmark & \checkmark \\
%\cline{2-9}
&Probab. loss&\checkmark&\checkmark&\checkmark&\checkmark& \checkmark & \checkmark & -\\
\cline{2-9}
&cSMOGN& \checkmark &\checkmark&\checkmark&\checkmark& *1 & \checkmark & - \\
\cline{2-9}
&crbSMOGN&-&-&-&-& - & - & \checkmark \\
\end{tabularx}
\caption{Combinations of all tested mitigation methods and relevance functions ("-" indicates, that the combination is not applicable).\\
*1: Skipped due to the similarity between kernel density relevance, DenseWeight and density-distance relevance (with uniform relevance distribution).\\
*2: Skipped due to the disadvantages of the interpolation-based relevance function in initial tests.}
\label{tab:combination}
\end{table}

The probabilistic loss function is not readily available in common machine learning libraries. Given the already large scope of the study and the poor performance of probabilistic loss for the MLP model, the implementation of both for XGBoosting tree and Random Forest is omitted.

We evaluate DenseLoss with a squared-error error, which makes it identical to SERA (as stated in Section \ref{sec:sera}). Therefore, we omit an evaluation of SERA.

In addition, we omit cost-sensitive learning based on BMC. BMC is only applicable under the assumption that the training data distribution is imbalanced, while the test data distribution is uniform. This is not the case in our evaluation and would therefore be an unfair comparison.

For density-distance relevance and density-ratio relevance, the probability density function associated with the domain priority $f_\text{p}$ is required in addition to the empirical distribution of the target variable $f_\text{x}$. If necessary, $f_\text{p}$ can be adjusted by the user to give more weight to certain areas of the domain. We use the mean squared error loss, which is based on a uniform cost assumption \citep{anand1993improved}. For better comparability with other relevance functions, we thus assume that there is no domain preference and the domain-priority distribution is uniform.

To ensure a principled evaluation of predictive performance under imbalance, the assessment procedure itself must avoid introducing overly restrictive assumptions about sample importance. The most unbiased evaluation would be obtained from a uniformly distributed test set covering the entire target domain. However, such a setting is rarely feasible in real-world regression problems, as extreme target regions typically contain only very few observations.

To address this issue, we evaluate performance using partition-based metrics (see Section \ref{sec:partitioned_loss_metrics}). Following \citet{steininger2021density} and \citet{camacho2024wsmoter}, we partition the target domain into five equidistant bins and rank them according to their sample frequency, ranging from very frequent to very rare. For each bin, the average loss over all contained samples is calculated.

While this discretization reduces dependence on explicit relevance formulations, it still implicitly defines importance through the target-domain partitioning itself. This may bias the evaluation in complex distributions, particularly in settings with multiple tails or heterogeneous density structures. Therefore, we extend the evaluation procedure by a second partitioning scheme using relevance values derived from the \nameref{sec:density_dist_relevance} function. Since the relevance values are bounded in the interval [0, 1], we discretize the relevance space into five equidistant bins ranging from least relevant to most relevant. Regression losses are then computed analogously within each relevance partition.

For brevity, we subsequently use the terms minority samples (or bin) to jointly refer to samples from the most relevant and very rare bins (or the bin directly) and majority samples (or bin) for the least relevant and very frequent ones, respectively.

To enable comparability across datasets with heterogeneous error scales, we apply a dataset-anchored normalization of the bin-wise losses in term of mean squared errors (MSE). Let $\mathrm{MSE}_{b}^{(r,d,c)}$ denote the bin-specific MSE for bin $b$, computed for repetition $r$, dataset $d$, and mitigation configuration $c$ (defined by mitigation method (with and without under-sampling), relevance function and model type).

First, for each tuple $(d, c)$, we aggregate the $50$ independent runs by computing the median of both the bin-wise MSE values and the overall (global) MSE. This yields a robust per-experiment summary that reduces sensitivity to stochastic variation across runs according to
\begin{equation} \label{eqn:median_error}
\widetilde{\mathrm{MSE}}_{b}^{(d,c)} = \mathrm{median}_{r}\left(\mathrm{MSE}_{b}^{(r,d,c)}\right), 
\qquad
\widetilde{\mathrm{MSE}}^{(d,c)} = \mathrm{median}_{r}\left(\mathrm{MSE}^{(r,d,c)}\right).
\end{equation}

Second, we construct a dataset-level reference scale for each dataset $d$ according to
\begin{equation}
\mathrm{MSE}_{\text{ref}}^{(d)} = \mathrm{median}_{c \in \mathcal{C}_{\text{no-mitigation}}}\left(\widetilde{\mathrm{MSE}}^{(d,c)}\right),
\end{equation}
where $\mathcal{C}_{\text{no-mitigation}}$ denotes configurations without any mitigation strategy across all model types.

Third, all bin-wise MSE values are normalized by this dataset-specific reference yielding a relative error ratio:
\begin{equation}
   \widetilde{\mathrm{RER}}_{b}^{(d,c)} = \frac{\widetilde{\mathrm{MSE}}_{b}^{(d,c)}}{\mathrm{MSE}_{\text{ref}}^{(d)}}. 
\end{equation}

The RER is dimensionless and represents multiplicative deviations from the dataset-level baseline. Having ruled out the impact of different target variable domain scales it allows for further aggregation across datasets to evaluate the performance of different mitigation strategies.

\begin{figure}
    \centering
    \includegraphics[width=1\textwidth]{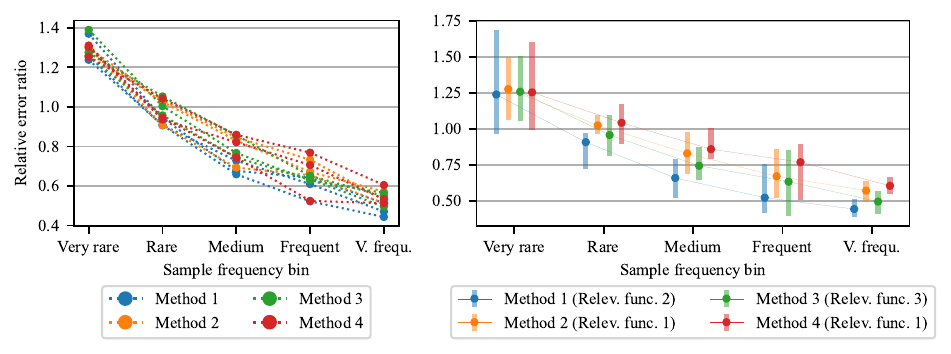}
    \caption{Exemplary bin-error plot using partitioning based on sample frequency. The figure demonstrates how model performance varies with respect to different sample frequency bins (very rare to very frequent) for multiple mitigation methods and relevance function configurations. \textbf{Left:} Median performance per configuration, showing results across frequency bins for each method--relevance function pair. Each curve corresponds to one configuration, and points represent median values over the datasets. \textbf{Right:} Performance of the best configuration per mitigation method (selected based on median performance in the very rare bin). Shaded error bars indicate the interquartile interval across datasets.}
    \label{fig:example_bin_error_plot}
\end{figure}

Figure \ref{fig:example_bin_error_plot} illustrates such an evaluation, whereby the bins $b$ are exemplary formed based on sample frequency. The example is based on synthetic data and is intended solely to illustrate the interpretation of the evaluation layout used throughout this study.

The left plot shows the median bin performances of all datasets according to
\begin{equation}
    \mathrm{RER}_{b}^{(c)} = \mathrm{median}_{d}\left(\widetilde{\mathrm{RER}}_{b}^{(d,c)} \right),
\end{equation}
where each curve corresponds to one mitigation configuration $c$.

The right plot shows only the best performing mitigation configuration per mitigation method. The shaded error bars indicate the interquartile interval across datasets, summarizing variability in performance. Specifically, for each mitigation method $m$, let $\mathcal{C}(m)$ denote the set of configurations sharing the same method but differing in relevance function, applied under-sampling and model type. We select the best configuration with respect to the minority (very rare or most relevant) bin according to
\begin{equation}
c^{*}(m) = \arg\min_{c \in \mathcal{C}(m)} \mathrm{median}_{d}\left(\widetilde{\mathrm{RER}}_{\text{minority}}^{(d,c)}\right).
\end{equation}

Using this selected configuration, the per-method aggregated performance for each bin is computed as
\begin{equation}
\mathrm{RER}_{b}^{(m)} = \mathrm{median}_{d}\left(\widetilde{\mathrm{RER}}_{b}^{(d,c^{*}(m))}\right),
\end{equation}
with variability across datasets summarized via the interquartile interval:
\begin{equation}
\mathrm{IQR interval}_{b}^{(m)} = \left[
Q_{0.25}\left(\widetilde{\mathrm{RER}}_{b}^{(d,c^{*}(m))}\right),
Q_{0.75}\left(\widetilde{\mathrm{RER}}_{b}^{(d,c^{*}(m))}\right)
\right].
\end{equation}

Similar plots can also be generated using binning of the sample relevance instead of the target-variable frequency. In the following, these visualizations are denoted as bin-error plots.

Assessing statistical significance is essential to determine whether observed differences in model performance across datasets are robust or merely due to random variation across repeated runs. This ensures that reported performance improvements are attributable to the applied mitigation strategies rather than stochastic effects.

To this end, statistical testing is performed directly on the run-level binned mean squared errors $\mathrm{MSE}_{b}^{(r,d,c)}$, without prior aggregation across repetitions $r$. For each dataset $d$, bin $b$, and mitigation configuration $c$, paired samples are constructed between a baseline configuration $c_{\text{base}}$ (no mitigation strategy) and a competing configuration $c$. The analysis is done for each model type individually, thus the model type is excluded from $c$.

Let $\{\mathrm{MSE}_{b}^{(r,d,c_{\text{base}})}\}_{r=1}^{R}$ and $\{\mathrm{MSE}_{b}^{(r,d,c)}\}_{r=1}^{R}$ denote the paired run-level error samples. The Wilcoxon signed-rank test is applied to these paired samples to assess whether their distributions differ significantly. The null hypothesis assumes no difference in median performance between both configurations, and statistical significance is evaluated at $\alpha = 0.05$.

A dataset $d$ is counted as a "significant win" for configuration $c$ in bin $b$ if the mean difference
\[
\Delta_{b}^{(d,c)} = \frac{1}{R}\sum_{r=1}^{R} \left(\mathrm{MSE}_{b}^{(r,d,c)} - \mathrm{MSE}_{b}^{(r,d,c_{\text{base}})}\right)
\]
is negative and the Wilcoxon test indicates statistical significance. Conversely, it is counted as a "significant loss" if the difference is positive and significant. Cases without statistical significance are recorded as normal wins or loses depending on the sign of $\Delta_{b}^{(d,c)}$. Datasets with fewer than five valid paired samples are excluded from the test to ensure robustness. This can happen if e.g., a bin does not contain any samples due to an imbalanced distribution of the data.

\begin{figure}
    \centering
    \includegraphics[width=1\textwidth]{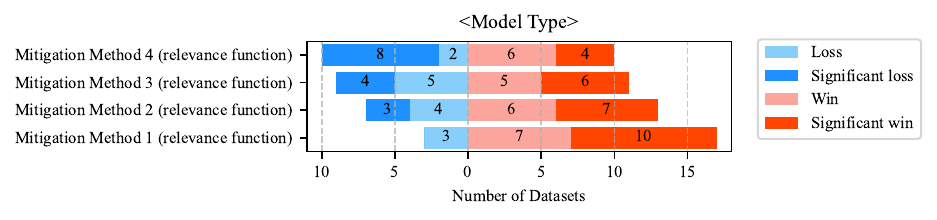}
    \caption{Exemplary bin-win plot. Number of datasets won and lost by models employing different mitigation strategies compared to a baseline (model trained without any mitigation strategy). The evaluation is performed individually for each specific sample frequency/relevance bins. Statistical significance is determined using the Wilcoxon signed-rank test with a significance threshold of 0.05. Significant wins and loses indicate where differences in model performance are statistically meaningful.}
    \label{fig:example_win_plot}
\end{figure}

Fig.~\ref{fig:example_win_plot} summarizes the number of datasets won and lost by each mitigation configuration relative to the baseline. This evaluation is performed independently for each bin $b$ defined over the sample frequency or relevance space. In the following, such visualizations are referred to as bin-win plots.

\subsubsection[Case study: synthetic data]{Case study with synthetic data}\label{sec:case_study_synthetic}
In this section, we evaluate the performance of combinations of relevance functions and mitigation methods on 10 synthetic datasets. Each dataset contains 1000 samples. For further information on the data generation process we refer to Appendix \ref{sec:appendix_data}.

During evaluation, we noticed that the interpolation-based relevance function is not able to reliably find relevance values for 2 of the 10 datasets (depending on the respective train/test random split). For one further dataset, the relevance function is unable to determine any relevance values at all. This is expressed by all the samples receiving the same binary relevance. The suspected cause is that the relevance function is only intended for distributions of normal-like shape with very marked trails due to the control point definition with boxplot statistics \citep{ribeiro2011utility}. 

\begin{figure}[htpb]
    \centering
    \includegraphics[width=1\textwidth]{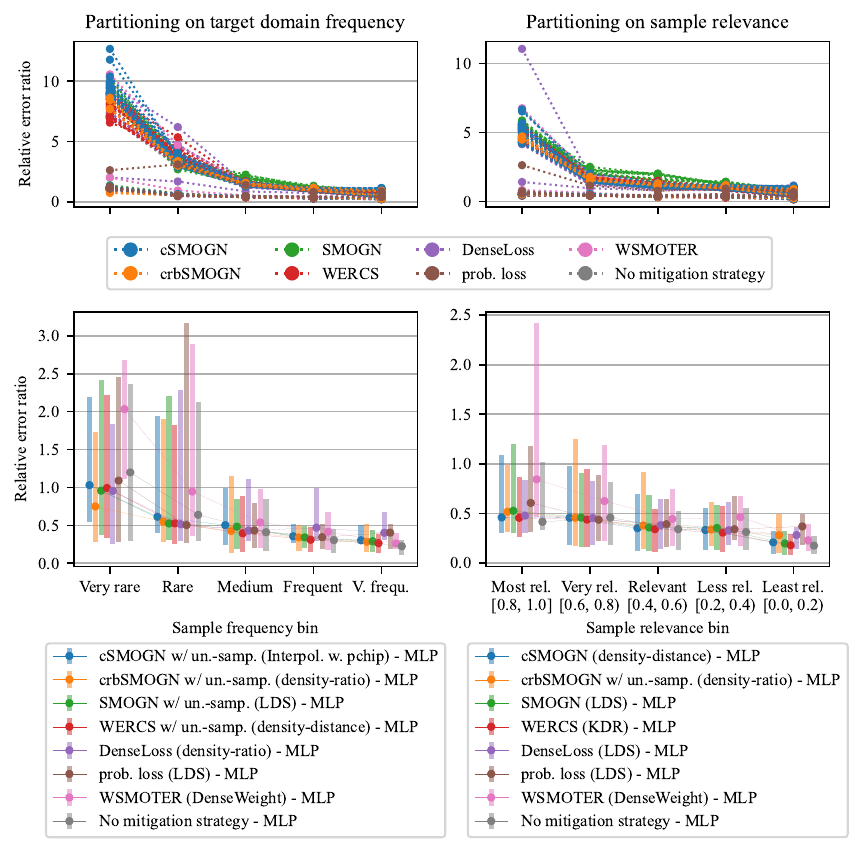}
    \caption{Bin-error plots of the synthetic datasets using either sample-frequency-based binning (left) or relevance-based binning (right) with density-distance relevance. The top subplots show of all tested combinations of relevance function, mitigation method and model type. The bottom subplots show only the best combination per mitigation method with respect to the very rare (most relevant) sample bin. Error bars indicate the interquartile interval across datasets.}
    \label{fig:bin_error_plot_synth}
\end{figure}

The bin-error plots across all synthetic datasets are shown in Fig. \ref{fig:bin_error_plot_synth}. The top part of the figure shows the results from all tested combinations of relevance function, mitigation method and model type. For a better overview, only the best combinations per mitigation method are shown in the bottom part.

In the top two plots, two clusters of lines can be identified considering the performance for minority samples. The higher cluster contains all Random Forest and XGBoosting trees models. The lower cluster only contains neural networks.

The difference between the model types can be explained by the fact that the synthetic datasets originate from regular (parametric) distributions. MLP neural networks are effective at capturing continuous, smooth, and well-structured relationships in the data and can thus efficiently learn the functional mapping between inputs and outputs. At the same time Random Forest and XGBoost model, based on decision trees, partition the feature space into discrete regions and do not benefit from the well-structured relationship.

One important observation should be emphasized here: even when using outlier-free synthetic data, all models exhibit a bias toward majority samples, regardless of whether a mitigation strategy is employed. However, as the results shown in Fig.~\ref{fig:bin_error_plot_synth} are aggregated across datasets, this trend should not be interpreted as true for every individual dataset. To assess the variability between datasets, the lower two plots show only the best-performing configuration (lowest minority bin error) for each mitigation method together with the interquartile interval across datasets.

Several observations can be made. First, the median relative error ratio generally decreases towards the majority bin. This trend is consistent across all displayed mitigation methods, indicating that the learning task becomes progressively easier in regions that are well represented in the training data.

Second, the interquartile intervals increase substantially towards minority bins. This indicates that performance in these regions varies considerably across datasets. A similar increase in variability can be observed for models trained without any mitigation strategy. Consequently, the larger spread cannot be attributed solely to the mitigation methods themselves but appears to be an inherent property of learning from sparsely represented regions of the target distribution.

Interestingly, the variability is considerably smaller when the bins are defined using relevance rather than sample frequency. This suggests that relevance-based binning yields more homogeneous groups of samples across datasets, whereas frequency-based bins may contain regions with substantially different prediction difficulty. One possible explanation is that relevance-based binning groups samples according to their estimated importance rather than their absolute occurrence frequency. For example, in a symmetric distribution such as a normal distribution, frequency-based binning would assign the left and right tails to separate rare and very rare bins according to their local sample densities. In contrast, relevance-based binning would group samples from both tails into the same highly relevant bin. Consequently, relevance-based bins may capture conceptually similar regions of the target distribution more consistently across datasets, thereby reducing variability in the aggregated performance estimates.

Using the data at hand, all best-performing configurations employ MLP neural networks and exhibit a decreasing error towards majority bins. Thus, even after applying data imbalance mitigation techniques, the resulting models remain biased. It is also noticeable that the best-performing WSMOTER configuration performs worse on average than the model trained without any mitigation strategy across all bins. While this observation is somewhat surprising, conclusions should be drawn with caution. The original WSMOTER study by \citet{camacho2024wsmoter} reports improvements on real-world datasets using a Wilcoxon signed-rank test. We therefore postpone a more rigorous assessment of this observation to Section~\ref{sec:case_study_real_world_data}.

Overall, most mitigation methods improve performance for minority, although this typically comes at the expense of reduced performance for majority. For these regions, models trained without any mitigation strategy generally achieve the lowest error. Whether these observations transfer to real-world datasets is investigated in the following section.

\subsubsection[Case study: real-world data]{Case study with real-world data}\label{sec:case_study_real_world_data}
In this section, the applicability of the data imbalance mitigation strategies to real-world datasets is evaluated. For this purpose, 42 datasets from different domains are used. For more information on the datasets we refer to the Appendix \ref{sec:appendix_data}.

As in the case study with synthetic data, the interpolation-based relevance function is partially not able to find reasonably distributed relevance values or cannot determine any relevance at all. In this case, 13 of the 42 datasets are affected. We therefore exclude the interpolation with pchip relevance function from the evaluation.

\begin{figure}[htpb]
    \centering
    \includegraphics[width=1\textwidth]{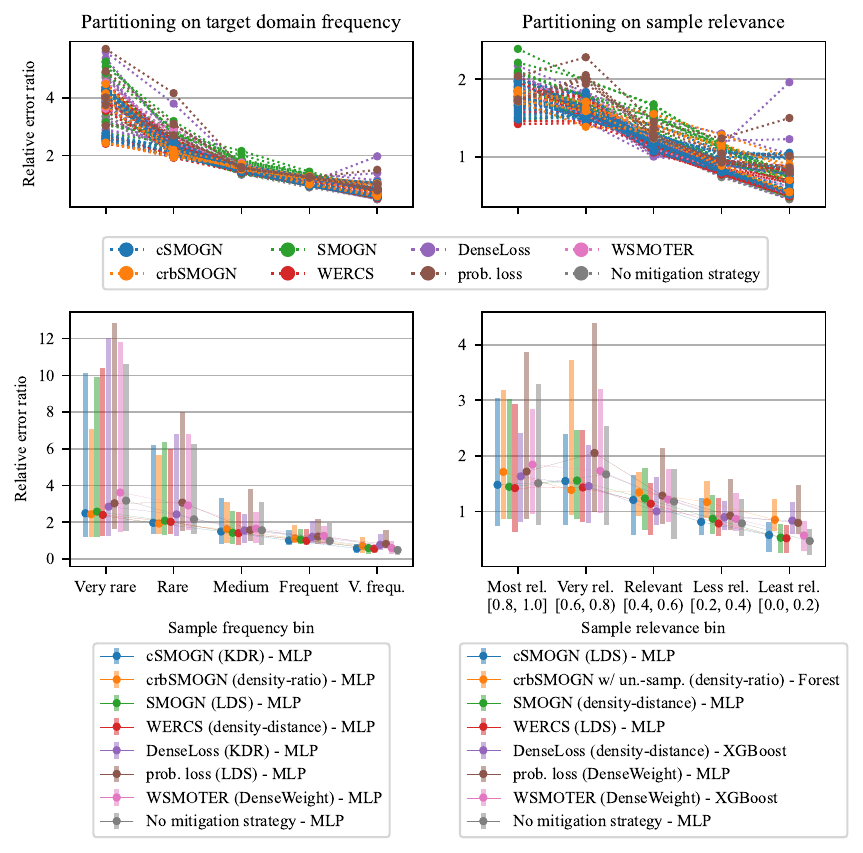}
        \caption{Bin-error plots of the real-world datasets using either sample-frequency-based binning (left) or relevance-based binning (right) with density-distance relevance. The top subplots show of all tested combinations of relevance function, mitigation method and model type. The bottom subplots show only the best combination per mitigation method with respect to the very rare (most relevant) sample bin. Error bars indicate the interquartile interval across datasets.}
    \label{fig:bin_error_plot_real}
\end{figure}

The bin-error plots across all real-world datasets is shown in Fig. \ref{fig:bin_error_plot_real}.

Considering the top two subplots showing the relative performance per bin for all combinations of mitigation method, relevance function and model types, we cannot observe clusters of lines as in the synthetic case. In addition, it can be seen that the performance is rather similar for medium frequent bins and diverges for very rare and very frequent bins. This effect is especially evident for DenseLoss, which is caused by assigning frequent samples a relevance of close to 0 and thus having no/very little impact on the training. Overall, the relative performance has wider spread between methods if sample relevance is considered for binning.

Considering the bottom subplots, which show only the best combination for each mitigation method, the decreasing interquartile interval between datasets can be seen towards majority bins, as in the synthetic data scenario. Also, the increasing error with sample rarity/relevance persists. The negative impact of WSMOTER for rare/relevant samples also persists. However, considering the size of the interquartile interval and thus the spread of performance between the datasets, no reliable statement can be made which methods are superior. 

To assess the robustness of the mitigation strategies across a diverse set of datasets, we additionally analyze the number of datasets won and lost relative to models trained without any mitigation strategy using the Wilcoxon signed-rank test.

\begin{figure}[htpb]
    \centering
    \includegraphics[width=1\textwidth]{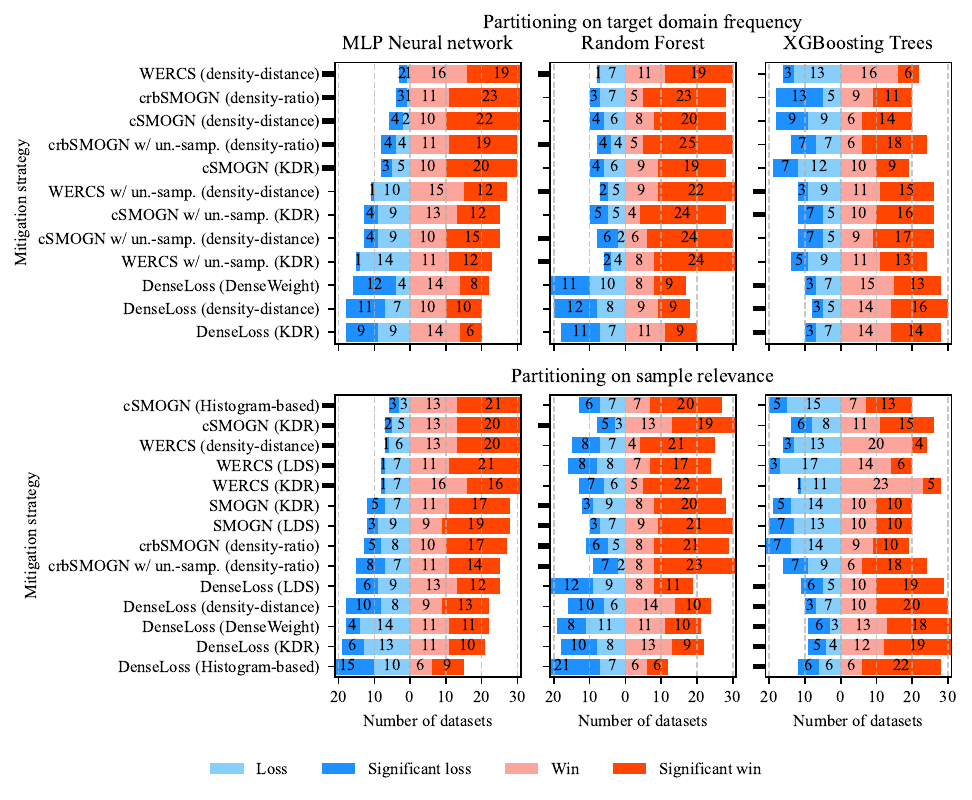}
    \caption{Bin-win plots of the best five mitigation strategies for each of the three model types. Performance is compared between models trained with the mitigation strategy to models trained without. Only the error of the very rare/most relevant sample bin is considered. The best five methods for each model type are indicated by bold tick marks.}
    \label{fig:bin_win_plot_real_top5}
\end{figure}

Fig.~\ref{fig:bin_win_plot_real_top5} presents the bin-win plots for the five best-performing mitigation strategies of each model type. The corresponding plots containing all evaluated strategies are provided in Appendix~\ref{sec:additional_figures}. The comparison focuses on the very rare and most relevant bins, as these represent the primary target regions of imbalance mitigation techniques.

It is important to note that the bin-win plots provide a one-vs.-one comparison between each mitigation strategy and the corresponding no-mitigation baseline. Consequently, the results quantify how often a strategy yields a statistically significant improvement over the baseline. They do not provide a direct comparison between different mitigation strategies. Therefore, a strategy that achieves a large number of wins should be interpreted as being effective relative to the baseline, but not necessarily as being superior to all competing mitigation approaches.

The results indicate that the effectiveness of a mitigation strategy depends strongly on the underlying model type. There is little overlap among the top-performing strategies of MLPs, Random Forests, and XGBoost models, suggesting that the interaction between mitigation strategy and learning algorithm is a dominant factor influencing performance.

For MLP models, sampling-based approaches consistently outperform cost-sensitive learning. Furthermore, for every sampling-based method, the pure over-sampler achieves a higher number of won datasets than its corresponding variant with under- and over-sampling. This trend is visible for both frequency-based and relevance-based binning and is further supported by the full bin-win plots shown in Appendix~\ref{sec:additional_figures}. 

These findings suggest that the removal of majority samples is generally detrimental for MLP models in the considered benchmark. In contrast, Random Forest models mostly benefit from the inclusion of under-sampling. For most mitigation strategies, the under-sampling variants achieve a larger number of won datasets than their over-sampling-only counterparts. 

A possible explanation is that tree induction is driven by frequency-weighted impurity reduction, such that densely populated majority regions disproportionately influence split selection. Under-sampling reduces this imbalance during tree construction and can therefore shift the split allocation toward minority regions, improving local resolution. In contrast, neural networks learn a global parametric function via gradient-based optimization, where reducing majority samples removes informative gradients that also shape shared representations across the entire input space. As a consequence, the model may lose also descriptive power in regions surrounding minority samples. Nevertheless, it should be noted that the under-sampling rate was not optimized in this study, and different parameter settings may lead to different conclusions.

XGBoost models exhibit a distinct behavior compared to both MLPs and Random Forests. Here, the cost-sensitive learning method DenseLoss achieves the largest number of wins against the no-mitigation baseline. A plausible explanation is that boosting algorithms already place increased emphasis on difficult-to-predict observations during training. Consequently, directly modifying the loss function may be more effective than altering the training data distribution.

Across all model types, the newly proposed relevance functions density-distance relevance and density-ratio relevance frequently appear among the best-performing ones. Similarly, the proposed mitigation methods cSMOGN and crbSMOGN consistently achieve competitive results and often rank among the strongest sampling-based methods.

Overall, the results highlight that the choice of mitigation strategy should not be considered independently of the learning algorithm. Instead, the effectiveness of a mitigation technique depends strongly on the interaction between the mitigation method and the underlying model architecture. At the same time, the bin-win plots only assess improvements relative to the absence of mitigation. An open question therefore remains whether the apparent differences between mitigation strategies are statistically significant when the strategies are compared directly against one another. To answer this question and identify the best-performing mitigation methods in an all-vs.-all comparison, we follow the non-parametric procedure proposed by \citet{demsar_statistical_2006} for comparing multiple learning algorithms across multiple datasets.

This procedure first applies the Friedman test to assess the null hypothesis that all methods perform equally. If this hypothesis is rejected, the Nemenyi post-hoc test is used to determine which pairs of methods differ significantly based on the critical difference (CD).

The Nemenyi test examines whether the average ranks of two methods across all datasets differ significantly, i.e., whether the observed difference in their mean ranks exceeds the critical difference, defined as
\begin{equation}
CD = q_\alpha \cdot \sqrt{\frac{k \cdot (k+1)}{6N}},
\end{equation}
where $k$ is the number of models, $N$ the number of datasets, and $q_\alpha$ the critical value of the Studentized range statistic. For all tests we use a significance limit of $\alpha = 0.05$.

Unlike the Wilcoxon signed-rank test used for the bin-win plots, which operates on the 50 paired repetitions within a single dataset, the Friedman and Nemenyi tests treat datasets as the statistical units of analysis. Consequently, the repeated training runs cannot be considered independent observations, as they correspond to repeated measurements of the same dataset--configuration combination. Treating these repetitions as separate observations would artificially inflate the sample size and violate the assumptions of the Friedman/Nemenyi procedure.

Therefore, each dataset contributes only a single performance value per configuration. To obtain this representative value, the 50 repetitions are aggregated by taking their median bin-wise error $\widetilde{\mathrm{MSE}}_{b}^{(d,c)}$ as defined in Eqn. \ref{eqn:median_error}. The resulting values are then ranked independently for each dataset before applying the Friedman and Nemenyi tests.

To isolate the effect of the mitigation method from that of the relevance function, we consider only the best-performing relevance function for each mitigation method. Thus, the Nemenyi test effectively evaluates whether the mean rank of one mitigation method differs significantly from that of another mitigation method when each is paired with its respective best-performing relevance function.

\begin{figure}[htpb]
    \centering
    \includegraphics[width=1\textwidth]{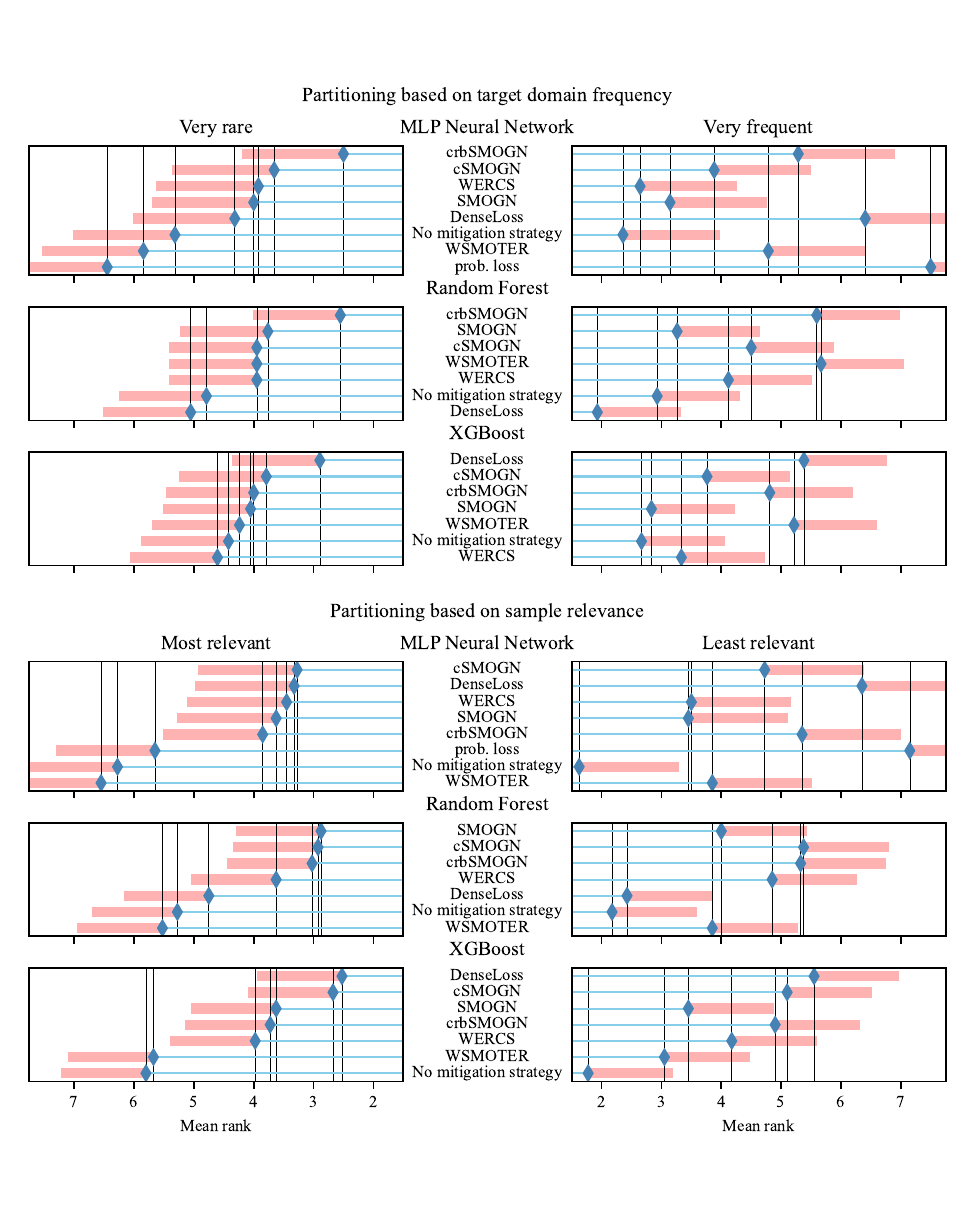}
    \caption{Results of the Nemenyi post-hoc test comparing different mitigation methods across datasets. For each mitigation strategy, the best-performing relevance function is selected. The red lines indicate the critical difference (CD). Two methods differ significantly if their average ranks differ by more than the CD.}
    \label{fig:critical_distance_real_combined}
\end{figure}

The previously shown bin-error plots indicated imbalance mitigation methods might cause a performance decrease for majority samples. Thus, we analyze performance for minority and majority bins. Fig.~\ref{fig:critical_distance_real_combined} presents the results of the Nemenyi post-hoc test for both frequency-based and relevance-based binning across all model types. 

Under both binning schemes, the results indicate that for XGBoost models DenseLoss achieves the highest average rank improvement for the minority bin compared to all other mitigation methods. This is consistent with the bin-win plot (Fig.\ref{fig:bin_win_plot_real_top5}). However, this improvement is accompanied by a pronounced degradation in performance for majority bins, consistent with the trends observed in the bin-error plot (Fig.~\ref{fig:bin_error_plot_real}).

For MLP and Random Forest models, crbSMOGN achieves the highest average ranks in the rare-sample regime. This effect is not explicitly visible in the bin-win analysis, which is possible given the different statistical perspectives of the two evaluations. The Wilcoxon-based analysis quantifies how consistently a method improves over the no-mitigation baseline across datasets, whereas the Nemenyi procedure operates on dataset-wise ranks and therefore captures relative performance differences between all competing methods. As a consequence, a method may achieve frequent but moderate improvements over the baseline (resulting in many significant wins), while still being outperformed by alternative mitigation strategies in direct comparisons, leading to lower average ranks. Conversely, a method with fewer but larger improvements may achieve superior ranks despite fewer significant wins.

A notable discrepancy between frequency-based and relevance-based evaluations is again observed for crbSMOGN. While the method performs strongly under frequency-based binning, its relative performance decreases under relevance-based binning. A plausible explanation is the mismatch between the relevance function used within crbSMOGN, which is based on density-ratio relevance, and the relevance function used for evaluation (density-distance relevance). Density-ratio relevance directly quantifies the degree of imbalance as a ratio, whereby density-distance relevance (as well as all other relevance functions) produces relevance scores bound between 0 and 1 and with different interpretations. Thus, the resulting bin assignments may differ substantially.

From a methodological perspective, the results suggest that crbSMOGN is particularly effective when the objective is to improve performance in rare-sample regimes, for MLP and Random Forest models, but causes a large performance decrease for frequent samples. In contrast, cSMOGN and WERCS provide a more balanced trade-off between rare and frequent sample performance under both frequency- and relevance-based evaluations. Across all settings, models without mitigation consistently achieve the best performance on majority samples, indicating that any form of mitigation introduces a trade-off between minority- and majority-region performance. This trade-off has been reported in prior work and remains evident even when using the newly proposed relevance functions and mitigation strategies. 

If extrapolated, this suggests that perfectly balanced performance across the entire target distribution may not be achievable by reweighting or resampling alone, but would require explicitly sacrificing performance in majority regions. An open question is therefore whether this limitation is fundamental to imbalance correction or whether alternative formulations could mitigate this effect without inducing a corresponding loss in majority-region performance.

\subsubsection[Case study: ensembles]{Case study with real-world data using ensembles}\label{sec:case_study_ensembles}
In this section, we investigate whether the deterioration in majority-sample performance observed for imbalance mitigation strategies can be alleviated through ensemble formation. To this end, we evaluate the ensemble approaches introduced in Sec.~\ref{sec:ensemble_formation}. Each ensemble consists of two models: a baseline model trained without any mitigation strategy and a model trained using a mitigation strategy. Their predictions are combined using either a mean-based ensemble, a weighted mean-based ensemble, or a threshold-based ensemble.

To assess the effect of ensemble formation, we compare the performance of each ensemble against a baseline MLP model trained without any mitigation strategy. Statistical significance is evaluated using the Wilcoxon signed-rank test, and the resulting numbers of wins and loses are aggregated across the 42 real-world datasets described in Appendix~\ref{sec:appendix_data}. The analysis focuses on both minority and majority regions of the target distribution using frequency-based and relevance-based binning.

The previous case studies demonstrated that MLP models exhibit strong performance on imbalanced regression tasks while also being particularly responsive to imbalance mitigation strategies. In contrast, Random Forest and XGBoost models are already ensemble methods by design. To avoid introducing nested ensemble structures, the present analysis is therefore restricted to MLP models. Furthermore, not all mitigation strategies are considered. The results in Sec.~\ref{sec:case_study_real_world_data} identified crbSMOGN, cSMOGN, and WERCS as the most promising approaches for reducing prediction bias in MLP models (see Fig.~\ref{fig:critical_distance_real_combined}). Consequently, the ensemble study focuses exclusively on these three mitigation methods.

\begin{figure}[htpb]
    \centering
    \includegraphics[width=1\textwidth]{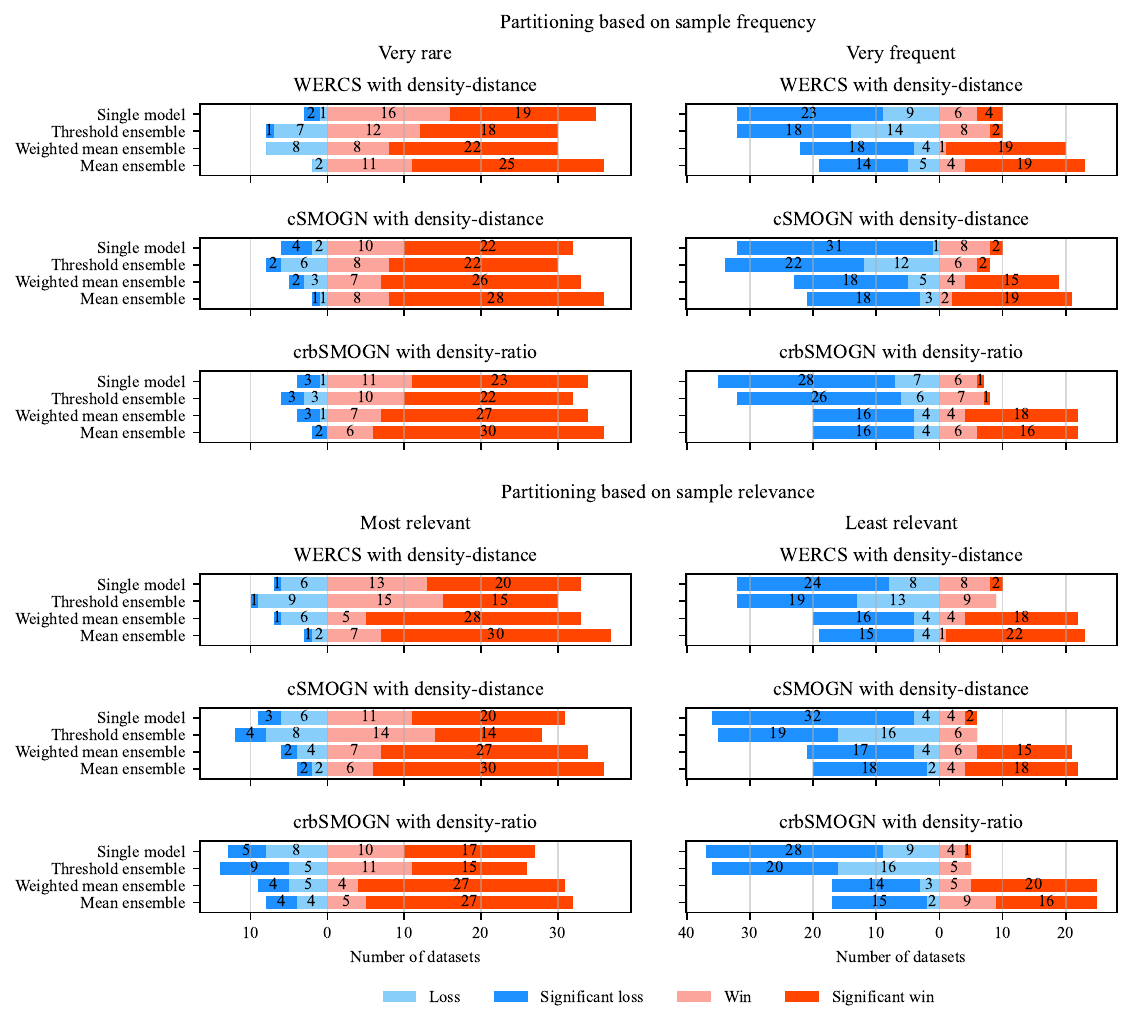}
    \caption{Bin-win plots for different ensemble formation techniques and mitigation methods evaluated on the 42 real-world datasets. Performance is compared against an MLP trained without any mitigation strategy using the Wilcoxon signed-rank test. The bars indicate the number of datasets for which a statistically significant win or loss is observed. For reference, the corresponding single-model mitigation strategies are included alongside the ensemble variants.}
    \label{fig:bin_win_plot_ensemble}
\end{figure}

Figure~\ref{fig:bin_win_plot_ensemble} presents the resulting bin-win plots for the different ensemble formation techniques and mitigation methods. For reference, the performance of the corresponding single-model mitigation strategy is also shown. 

The left column shows the results for minority samples. Across all three mitigation methods, the mean-based ensemble increases both the number of statistically significant wins and the overall number of wins relative to the no-mitigation baseline when compared to the corresponding single-model approach. Consequently, the number of lost datasets is reduced. Among the evaluated ensemble formation techniques, the mean-based ensemble consistently achieves the strongest performance. In contrast, threshold-based ensembles generally perform worse than the corresponding single-model mitigation strategy. The weighted mean-based ensemble produces more mixed results, but does not outperform the mean-based ensemble in any setting.

The right column shows the results for majority samples. As demonstrated in the previous sections, imbalance mitigation strategies frequently reduce predictive performance in majority regions of the target distribution. This behavior is clearly visible for the single-model mitigation approaches, which lose against the no-mitigation baseline on most datasets. Ensemble formation substantially mitigates this effect. Both mean-based and weighted mean-based ensembles considerably reduce the number of loses and frequently achieve a near-balanced ratio between won and lost datasets. These results suggest that ensemble formation not only preserves much of the minority-sample improvement obtained through the mitigation strategy but can also partially recover the performance degradation observed for majority samples.

Nevertheless, the approximately balanced number of wins and loses in the majority bins indicates that the effectiveness of ensemble formation remains strongly dataset-dependent. While ensembles often reduce the negative side effects of mitigation strategies, they do not eliminate them entirely.

Overall, the mean-based ensemble emerges as the most effective of the three evaluated ensemble formation approaches. The results suggest that combining a mitigated model with an unmitigated baseline model can substantially improve minority-sample performance while simultaneously reducing the loss in majority-sample performance that typically accompanies imbalance mitigation.

Although these findings are encouraging, they should be interpreted as an initial exploratory study. First, the Wilcoxon signed-rank test only indicates whether performance differences are statistically significant and does not quantify their magnitude. Second, the present evaluation is restricted to ensembles consisting of two models. More complex ensemble designs, including larger ensembles and alternative ensemble formation techniques such as stacking, may yield further improvements. Investigating such approaches constitutes an interesting direction for future work.

\section{Discussion} \label{sec:discussion}
In the previous sections, new relevance functions and mitigation methods were presented and evaluated. However, the approach chosen here is by no means beyond doubt, which is why we put some points up for discussion. At the end of the discussion, we will also look at some application recommendations for future users.

First of all, it should be noted that none of the relevance functions really reflects how important a sample is for the modeling task in question. All presented relevance functions transfer the principle of empirical frequency of data points into a relevance value, which, depending on the model type, does not necessarily have to be decisive for model performance. Only density-distance relevance and density-ratio relevance provide the flexibility to combine empirical frequency and domain priority, which makes it possible to incorporate the importance of the sample for the modeling task. However, this may too be detached from the difficulty the model has in correctly predicting the sample. The perfect relevance function would therefore give a high relevance to samples that are "hard-to-predict" rather than based on sample rarity. Determining in advance which sample is "hard to predict" is model- and application-dependent and thus a very difficult task, which is why we restrict ourselves to the empirical frequency and domain priority.

For all kernel density-based relevance functions, we relied on Silverman’s rule of thumb for bandwidth estimation. However, this choice is not ideal, as the Silverman bandwidth is intended for approximately normal distributions and tends to over-smooth otherwise. Nonetheless, Silverman’s estimator remains the most commonly applied choice \cite{steininger2021density, camacho2024wsmoter, wibbeke2024quantification}, while systematic studies of alternatives are lacking. For these reasons, we opted for Silverman’s estimator, while acknowledging its potential limitations.

In the benchmark, domain priority was assumed to be uniform for density-distance and density-ratio relevance. Theoretically, it would have been possible to use arbitrary domain-priority distribution, however, then the comparability with relevance functions without domain priority would not be possible. In addition, expert knowledge of the domain is required for each dataset, which is not feasible to archive in a quantitative analysis. 

Due to the assumption of uniform domain priority, density-distance relevance, kernel density relevance and DenseWeight differ only slightly. This is also reflected in the small performance differences in the evaluation, which can be largely attributed to statistical noise. The newly presented density-distance relevance should therefore by no means be seen as competition to kernel density relevance and DenseWeight, but rather as a generalization with possibility for expansion by including domain priority.

Considering under-sampling, our analysis suggests that under-sampling, degrades model performance for MLP models and improves performance for Random Forests. However, this conclusion is constrained by the specific sampling strategies and fixed under-sampling rates (e.g., WERCS at 0.5) employed in our experiments. While we adopted the hyperparameters proposed by the original authors of these methods to ensure reproducibility, we recognize that optimizing under-sampling rates could yield different outcomes. The same holds true for the optimization of model parameters and over-sampling rate. Such an optimization, however, would significantly expand the computational and analytical scope of this benchmark as it would have to be done for each dataset individually. Future work could explore such optimization to better assess their potential in imbalanced regression tasks in dependence of the specific application and dataset. When applying imbalance mitigation to a specific use case, users should definitely try out different sampling rates.

Finally, we would like to address the validity of the results. Even if many of the mitigation strategies have a positive influence on the models of most regression tasks, this is by no means always the case. The evaluation approach chosen in this article is quantitative. We have aimed to put it on statistically sound footing by selecting a large number of datasets from various application domains, repeating the modeling 50 times and using established statistical tests to test for statistical significance. However, outliers cannot be ruled out and it must be expected that the chosen mitigation strategy may not achieve any improvement for a specific dataset.

Furthermore, for the time being, no reliable statements can be made about which properties a dataset or a modeling task must fulfill for the use of a mitigation strategy to be successful in general.

Based on the empirical findings, several practical recommendations can be derived for model bias mitigation in imbalanced regression. The following guidelines provide a starting point for selecting an appropriate method depending on the predictive objective, computational constraints, and application requirements:

Although crbSMOGN consistently achieved the strongest mitigation of imbalance-induced bias for MLPs, WERCS deserves particular attention. Originally introduced with interpolation-based relevance using pchip \citep{branco2019pre}, WERCS remains highly competitive when combined with kernel-density or density-distance relevance functions. Despite its conceptual simplicity, it offers effective bias mitigation at lower computational cost. Similarly, the ensemble evaluation demonstrated that simple aggregation strategies are often highly effective. Among the threshold-, weighted mean-, and mean-based ensembles, the mean-based variant consistently provided the strongest overall performance across the full target domain.

However, the effectiveness of mitigation strategies depends strongly on the underlying model type. Furthermore, both the under-sampling and over-sampling rates constitute important hyperparameters and should therefore be included in the hyperparameter optimization process. While the choice of relevance function influences performance, its impact is typically smaller than that of the mitigation strategy itself. Finally, performance varies substantially across datasets. For some datasets, sampling yields only marginal improvements, whereas for others, substantial gains can be achieved. Consequently, no universally optimal strategy exists, and the most appropriate solution must be determined individually for each application.

Nevertheless, when looking at the results of the provided benchmark some approaches that will most likely be successful can be identified: In general, we recommend using density-distance or density-ratio relevance. When a single model should be used and predictive performance on minority samples is the primary objective, an MLP trained with crbSMOGN is most likely a good choice. If performance for majority samples is also of importance cSMOGN might be the better call, with WERCS being a more computation cost-aware alternative. 

However, if model performance on both ends of the relevance spectrum is of major concern, we highly recommend to use a mean-based ensemble of MLP models. Here, crbSMOGN showed the best results in our study. When computational resources are limited, implementation simplicity is required, or optimal performance on rare samples is not essential, a mean-based ensemble of MLPs trained with WERCS and density-distance relevance provides an attractive alternative.

\section{Outlook} \label{sec:outlook}
In this section, we address open questions and problems that have arisen from our work.

At present, it is not possible to say in advance whether a mitigation strategy will be successful for the specific modeling task. During the evaluation is has become evident that not all mitigation strategies fail with the same datasets. It would therefore be interesting to investigate which specific characteristics datasets or modeling tasks must have for mitigation strategies to work particularly effectively or not. This would have the advantage that a priori of applying a mitigation approach a statement about the applicability could be made. This probably also depends on the distribution and importance of the individual features, which was not part of this work.

In addition, the questions arise as to how expert knowledge can be obtained to determine the domain priority and which of the two relevance functions allowing domain priority (density-distance and density-ratio relevance) is more powerful when including it.

Current relevance functions assess sample relevance exclusively from the marginal distribution of the target variable. However, sample rarity may also stem from sparsely populated regions of the input space. Extending relevance estimation to account for joint input--target distributions is a natural direction, for example by replacing marginal target densities with joint density estimates, as partially explored by \citet{yang2021delving}. While conceptually appealing, this substantially increases computational and statistical complexity, particularly when comparing joint distributions to domain-priority distributions.

So far, all density-based relevance functions used the Silverman's rule of thumb to assess the bandwidth for the kernel and we adhered to it for comparability and simplicity. However, the Silverman bandwidth is non-optimal for non-normal distributions. In the future, we therefore recommend evaluating whether other bandwidth estimators, like the ISJ bandwidth, are more suitable.

The distribution of the modeling error over the target value domain is far from uniform even when using mitigation methods. Moreover, most sampling-based mitigation methods are based on the same "building blocks", such as interpolation between data points using SMOTE or the use of Gaussian noise. The development of new building blocks could bring substantial progress and allow the formation of new mitigation methods. Creating new samples with data-driven generative methods could come into play here. For example, \citet{liu_research_2024} and \citep{belhaouari_oversampling_2024} investigate this direction using autoencoders. However, the requirements on dataset size and potential bias in the sample generation process are limiting factors to be overcome.

In previous work by \citet{wibbeke2024quantification} the mean imbalance ratio mIR was proposed to quantify data imbalance. However, here it was shown that imbalance mitigation strategies have different effects depending on the model type. For example, although under-sampling can lead to a lower mIR, it has a negative effect on the performance of MLP models whereby having positive impact on Random Forests. A possible remedy could be to quantify data imbalance using the imbalanced sample percentage of under-represented samples ISP$_{<t_u}$ \citep{wibbeke2024quantification}, instead of the mIR. The ISP$_{<t_u}$ only includes data points that are too rare and thus accounting for the fact that under-sampling might be harmful. 

From a general perspective, however, we suggest separating the concept of purely empirical data imbalance (detectable by the mIR or ISP) from the non-uniform error distribution of the model over the target domain (e.g., model error imbalance). However, a way to quantify the model error imbalance requires further elaboration and investigation.

\section{Conclusion} \label{sec:conclusion}
Data imbalance is a pervasive challenge in regression tasks, impacting the performance and reliability of predictive models to favor majority classes, resulting in a biased performance towards frequently occurring data points. In this article, various approaches to avoid this model bias based on cost-sensitive learning or sampling were presented. 

It was shown that every mitigation strategy consists of two components. The relevance function, which assigns a value of importance to the data points, and the mitigation method, which uses the relevance for sampling or cost-sensitive learning. Together, the combination of relevance function and mitigation method results in a mitigation strategy to reduce the model bias. Based on this separation, the most common relevance functions and mitigation methods were identified and compared. Additionally, we presented two new approaches for each.

The new relevance functions density-distance relevance and density-ratio relevance are based on the difference between the empirical frequency of the data points and any domain specific priority of the target variable, making it possible to combine the two for the first time. They offer a generalization of established approaches, which only consider either the empirical frequency or the domain priority. The newly proposed mitigation methods cSMOGN and crbSMOGN are based on established sampling methods, but offer advantages in that the user does not have to decide on a relevance threshold, as well as better sample selection process by discretization of the search space in the dataset. 

In the evaluation, various combinations of relevance functions and mitigation strategies were examined and the performance compared in a quantitative study. For this purpose, we used MLP neural networks, XGBoosting trees, and Random Forest models. To be able to make statistically reliable statements, 10 synthetic and 42 real-world datasets were used and the results were tested for statistical significance using the Wilcoxon signed-rank and Nemenyi test. 

It has been shown that the newly introduced mitigation method crbSMOGN with the associated ratio-based relevance function offers the best performance for rare samples (MLP model). However, the effectiveness of each mitigation strategy largely depends on the model type used.

The results reveal a persistent trade-off between minority- and majority-sample performance. Mitigation strategies that improve predictive performance in rare or highly relevant regions of the target distribution generally deteriorate performance in well-represented regions. Furthermore, the strongest gains for minority samples are often associated with the largest losses for majority samples.

For alleviating this trade-off, we introduced ensembles. By combining a model trained with a mitigation strategy and a model trained without mitigation, it is possible to retain the benefits of imbalance mitigation for minority samples while reducing its adverse effects on majority samples. Among the investigated approaches, mean-based ensembles consistently achieved the most favorable balance between these competing objectives.

These findings suggest that ensemble formation constitutes a viable strategy for improving performance on imbalanced regression problems without fully sacrificing predictive accuracy in majority regions of the target space. Consequently, the combination of imbalance mitigation and ensemble learning appears to be a promising direction for future research.

The main take-away points of this work are:
\begin{itemize}
    \item We introduced dual-source relevance functions as a generalization of already existing relevance functions, allowing for the inclusion of the empirical frequency of the data and any domain specific priority.
    \item We introduced new sampling method cSMOGN and crbSMOGN.
    \item The effectiveness of mitigation strategies largely depends on the model type, e.g. cost-sensitive learning approaches are better suited for XGBoosting tree models, whereby MLPs work best with over-sampling while Random Forests benefit from under-sampling.
    \item For MLP models...
    \begin{itemize}
        \item ...under-sampling causes a higher model error for frequent samples, while not increasing the performance for rare samples. Thus, we recommend to only use over-sampling.
        \item ...the choice of relevance function has less influence than the choice of mitigation method.
    \end{itemize}
    \item While decreasing the model error for rare samples, mitigation methods increase the error for frequent samples.
    \begin{itemize}
        \item The effect increases with mitigation strategy efficiency: the better the performance for rare samples the worse the performance for frequent ones.
        \item The trade-off can be reduced when constructing an ensemble. 
    \end{itemize}
    \item The recommend the usage of either crbSMOGN with density-ratio relevance or cSMOGN with density-distance relevance for sampling, with WERCS being a performative simpler alternative with less computational costs.
\end{itemize}

As a future perspective, we suggest to separate the concept of data imbalance from the non-uniform distribution of the model error over the target domain (e.g., model error imbalance). However, a way to quantify the model error imbalance requires further elaboration and investigation.

\section*{Acknowledgment}
This research was supported by the German Federal Ministry of Education and Research (Bundesministerium für Bildung und Forschung, BMBF) within the H$_2$Giga Initiative under grant number 03HY122F.

\bibliographystyle{unsrtnat}
\bibliography{references, dataset_bib}  %%% Uncomment this line and comment out the ``thebibliography'' section below to use the external .bib file (using bibtex) .

\appendix

\section{Selection of hyperparameters}
\label{sec:appendix_hyperparameter}
This section lists the various hyper-parameters used for model training, relevance functions and mitigation methods.

As hyperparameters for the neural network, a batch size of 16, a layer size of 128 with 2 hidden layers and an Adam optimizer with a learning rate of 0.001 is used. The training was run for 1000 epochs with early stopping. Other batch and layer sized where tested, but showed no significant difference. The extreme gradient boosting trees\footnote{\url{https://xgboost.readthedocs.io}} and Random Forests\footnote{\url{https://scikit-learn.org}} are initialized with the recommend standard hyperparameters respective library.

The Isolation Forest outlier detection was used with a probability threshold of 0.8 \citep{liu2012isolation}.

The Silverman's rule of thumb to estimate the bandwidth of the kernel was used for all relevance functions requiring kernel density estimation \cite{SilvermanB}.

For all mitigation methods allowing for the generation of synthetic samples using gaussian noise the magnitude of noise is 0.01 of the standard deviation of the respective variable.

The hyperparameters of the individual relevance functions and mitigation methods can be seen in Table \ref{tab:hyperparameter}.
\begin{table}[htbp]
    \centering
    \renewcommand{\arraystretch}{1.2}
    \caption{Used hyperparameters of the individual relevance functions and mitigation methods.}
    \begin{tabular}{|l|l|l|}
        \cline{1-3}
        \textbf{Method name} & \textbf{Hyperparameter} & \textbf{Value} \\
        \cline{1-3}
         \multirow{3}{*}{Interpolation with control points}& method & balanced \\
         \cline{2-3}
          & extremes & both \\
          \cline{2-3}
          & coefficients & 1.5 \\
         \cline{1-3}
         Histogram-based relevance & number of bins & 10 \\
          \cline{1-3}
         \multirow{3}{*}{Label distribution smoothing} & number of bins & 50\\
         \cline{2-3}
         & kernel width (bins) & 5 \\
         \cline{2-3}
         & kernel variance (bins) & 4\\
         \cline{1-3}
         DenseWeight & alpha & 1\\
         \cline{1-3}
         \multirow{3}{*}{SMOGN} & kNN & 5\\
         \cline{2-3}
         & sampling method & balanced \\
         \cline{2-3}
         & relevance threshold & 0.8\\
         \cline{1-3}
         \multirow{2}{*}{WERCS} & under-sampling rate & 0.5\\
         \cline{2-3}
         & over-sampling rate & 0.5\\
         \cline{1-3}
         \multirow{2}{*}{WSMOTER} & kNN & 10\\
         \cline{2-3}
         & over-sampling ratio & 3\\
         \cline{1-3}
         \multirow{2}{*}{cSMOGN} & number of bins for discretization & 10\\
         \cline{2-3}
         & allowed bin for deviation & 1 \\
         \cline{1-3}
         \multirow{4}{*}{crbSMOGN} & number of bins for discretization & 10\\
         \cline{2-3}
         & allowed bins for deviation & 1\\
         \cline{2-3}
         & under-sampling rate & 0.5 \\
        \cline{2-3}
         & over-sampling rate & 0.5 \\
         \cline{1-3}
    \end{tabular}
    \label{tab:hyperparameter}
\end{table}

\section{Datasets}
\label{sec:appendix_data}
The list of the real-world datasets used is provided in Table \ref{tab:datasets}. The list of all synthetic datasets used is provided in Table \ref{tab:synthetic_data}. Some of the synthetic datasets are created using the sklearn library \citep{scikit-learn}. For further information on the data generation process we refer to the provided code on GitHub\footnote{\url{github link to be added}}.

The case study on synthetic data is conducted using both clean and noisy data. For noise, each feature is assigned a normally distributed noise with a standard deviation of 3\% of the standard deviation of said feature. However, during the evaluation, we found that the results show no major difference between noisy and clean data. Thus, since noisy data corresponds more to the reality of measured data, only the noisy results are included in the article.

\begin{table*}[htbp] % the * indicates a two-column stretch for 2-col layouts
\caption{List of used datasets.}
\centering
\begin{tabular}{rlrrl}
No. & Name & Features & Samples & Ref. \\
\hline
1 & energy-efficiency & 8 & 768 & \citep{misc_energy_efficiency_242}\\
2 & forest-fires & 12 & 517 & \citep{data_forest_fires_162} \\
3 & optical-interconnection-network & 9 & 640 & \citep{misc_optical_interconnection_network__449} \\
4 & concrete-compressive-strength & 8 & 1030 & \citep{data_concrete_compressive_strength_165} \\
5 & servo & 4 & 167 & \citep{data_servo_87} \\
6 & combined-cycle-power-plant & 4 & 9568 & \citep{data_combined_cycle_power_plant_294} \\
7 & grid-stability & 12 & 10000 & \citep{data_electrical_grid_stability_simulated_data__471} \\
8 & superconductivity-data & 81 & 21263 & \citep{misc_superconductivty_data_464} \\
9 & synchronous-machine & 4 & 557 & \citep{data_synchronous_machine_data_set_607} \\
10 & auction-verification & 7 & 2043 & \citep{ordoni2022analyzing}\\
11 & airfoil-self-noise & 5 & 1503 & \citep{data_airfoil_self-noise_291} \\
12 & concrete-slump-test & 9 & 103 & \citep{misc_concrete_slump_test_182} \\
13 & traffic-behavior & 17 & 135 & \citep{misc_behavior_of_the_urban_traffic_of_the_city_of_sao_paulo_in_brazil_483} \\
14 & yacht-hydrodynamics & 6 & 308 & \citep{data_yacht_hydrodynamics_243} \\
15 & fish-toxicity & 6 & 908 & \citep{cassotti2015similarity} \\
16 & wave-energy-perth-49 & 149 & 63600 & \citep{neshat2020optimisation} \\
17 & aquatic-toxicity & 8 & 546 & \citep{cassotti2014prediction} \\
18 & steel-industry & 8 & 35040 & \citep{VE2020EfficientEC} \\
19 & computer-hardware & 8 & 209 & \citep{data_computer_hardware_29} \\
20 & abalone & 7 & 4177 & \citep{data_abalone} \\
21 & age-prediction & 9 & 6287 & \citep{Dinh2019ADA} \\
22 & parkinson & 18 & 5875 & \citep{tsanas2009accurate} \\
23 & winequality-white & 11 & 4898 & \citep{Cortez2009ModelingWP} \\
24 & facebook-metrics & 22 & 500 & \citep{data_facebook_metrics_368} \\
25 & ailerons & 40 & 13750 & \citep{data_torgo} \\
26 & anacalt & 7 & 4052 & \citep{data_keel} \\
27 & autoprice & 15 & 159 & \citep{data_torgo} \\
28 & bank32 & 32 & 8193 & \citep{data_delve} \\
29 & bank8 & 8 & 8193 & \citep{data_delve} \\
30 & baseball & 17 & 337 & \citep{data_baseball} \\
31 & boston-housing & 13 & 506 & \citep{data_delve} \\
32 & california & 8 & 20640 & \citep{data_torgo} \\
33 & deltaail & 5 & 7129 & \citep{data_torgo} \\
34 & deltaelv & 6 & 9517 & \citep{data_torgo} \\
35 & ele-1 & 2 & 495 & \citep{data_keel} \\
36 & ele-2 & 4 & 1056 & \citep{data_keel} \\
37 & house16 & 16 & 22784 & \citep{data_delve} \\
38 & laser-generated & 4 & 993 & \citep{data_keel} \\
39 & mortage & 15 & 1049 & \citep{data_keel} \\
40 & puma32 & 32 & 8192 & \citep{data_delve} \\
41 & strikes & 6 & 625 & \citep{data_strikes} \\
42 & treasury & 15 & 1049 & \citep{data_keel} \\
\end{tabular}
    \label{tab:datasets}
\end{table*}

\begin{table}[htpb]
\centering
\renewcommand{\arraystretch}{1.5}
\begin{tabularx}{\textwidth}{rlrX}
No. & Name & Features & Description \\
\hline
1 & Euclidean distance & 2 &  $y= \sqrt{a^2 + b^2}$\\
2 & Nernst equation & 6 &  $y= \frac{R \cdot T}{z \cdot F} \cdot \text{log}\left(\frac{a_1}{a_2}\right)$\\
3 & Stribeck translational friction & 6 &  $y= \mu_1 \cdot F + \left(\mu_2 \cdot F + \mu_1 \cdot F \right)\cdot \text{exp}\left(- |\frac{\vartheta_1}{\vartheta_2}|^\delta\right)$\\
4 & Inverse tangent & 1 &  $y= \text{arctan}(x)$\\
5 & Multilayer perceptron model & 4 & Random initialization of a MLP model with gaussian weights.\\
6 & Random linear model & 3 &  Dataset using the \verb|make_regression| function of sklearn*.\\
7 & Sparse uncorrelated & 4 &  Dataset using the \verb|make_sparse_uncorrelated| function of sklearn*.\\
8 & Friedman1 & 5 &  Dataset using the \verb|make_friedman1| function of sklearn*.\\
9 & Friedman2 & 4 &  Dataset using the \verb|make_friedman2| function of sklearn*.\\
10 & Friedman3 & 4 &  Dataset using the \verb|make_friedman2| function of sklearn*.\\
\end{tabularx}
    \caption{List of generated synthetic datasets. *: \url{https://scikit-learn.org}}
    \label{tab:synthetic_data}
\end{table}

\section{Additional bin-win plots} \label{sec:additional_figures}
This appendix provides the complete bin-win plots for all evaluated mitigation strategies and model types related to \ref{sec:case_study_real_world_data}~\nameref{sec:case_study_real_world_data}. While the main text shows only the best-performing strategies per model type, the full plots allow a more comprehensive assessment and confirm the results stated in the main body of the article. 

\begin{figure}[htpb]
    \centering
    \includegraphics[width=\textwidth]{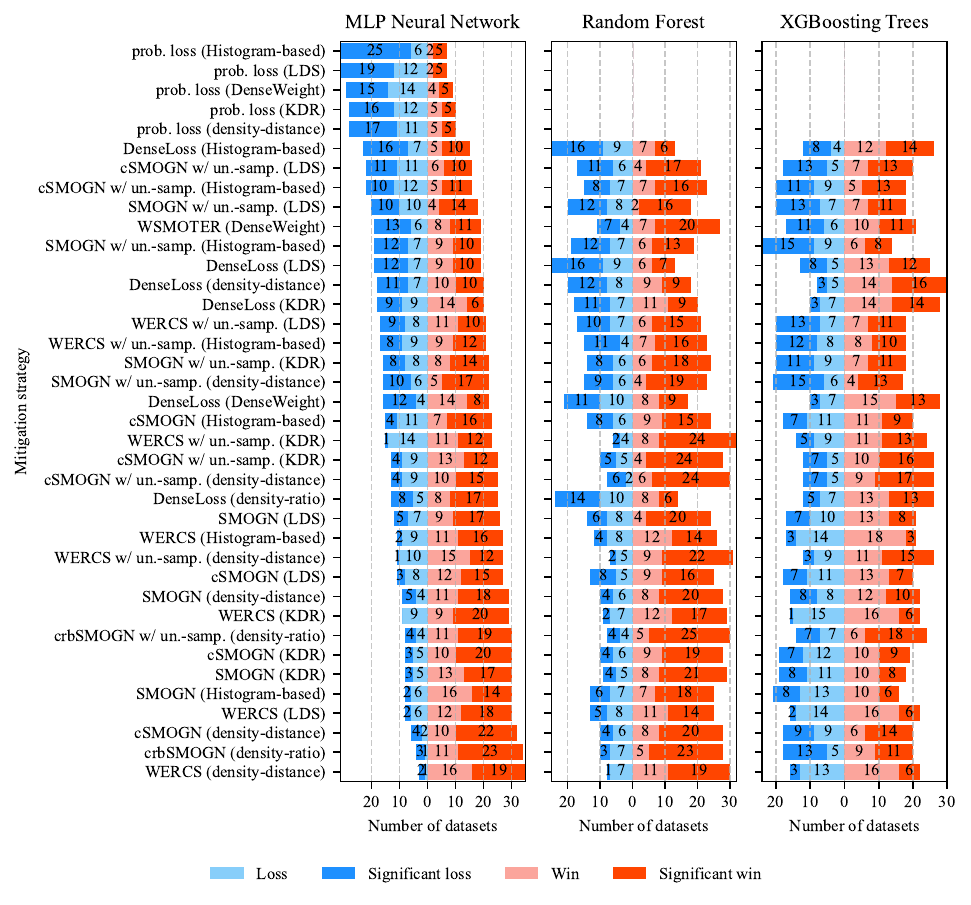}
    \caption{Full bin-win plots for all evaluated mitigation strategies and model types using frequency-based binning. Each mitigation strategy is compared against a model of the same type trained without any mitigation strategy. The plots show the number of datasets for which a (statistically significant) win or loss is observed in the very rare sample bin according to the Wilcoxon signed-rank test.}
    \label{fig:real_bins_full_frequency}
\end{figure}

\begin{figure}[htpb]
    \centering
    \includegraphics[width=\textwidth]{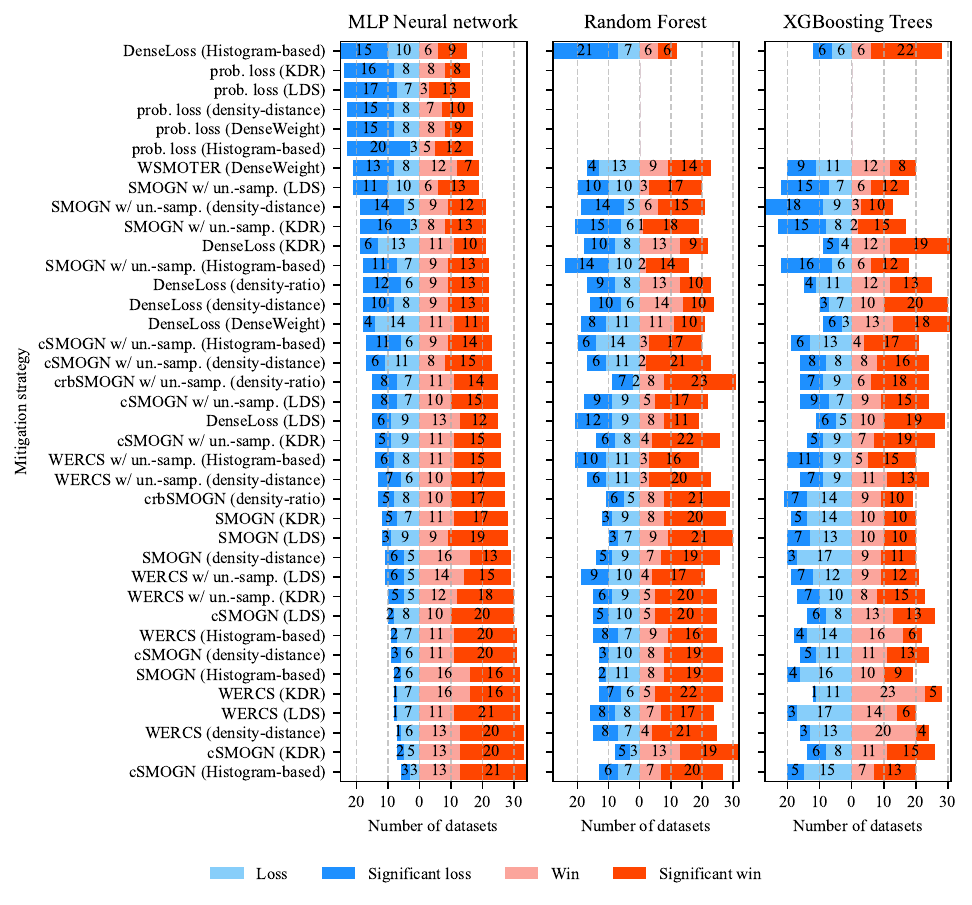}
    \caption{Full bin-win plots for all evaluated mitigation strategies and model types using relevance-based binning with density-distance relevance. Each mitigation strategy is compared against a model of the same type trained without any mitigation strategy. The plots show the number of datasets for which a (statistically significant) win or loss is observed in the most relevant sample bin according to the Wilcoxon signed-rank test.}
    \label{fig:real_bins_full_relevance}
\end{figure}

Figures~\ref{fig:real_bins_full_frequency} and \ref{fig:real_bins_full_relevance} summarize the results of the Wilcoxon signed-rank analysis across all evaluated datasets. Figure~\ref{fig:real_bins_full_frequency} uses bins defined by target-variable frequency, whereas Fig.~\ref{fig:real_bins_full_relevance} uses bins defined by sample relevance. In both cases, each mitigation strategy is compared against a model of the same type trained without any mitigation strategy. Consequently, the plots quantify the improvement relative to the no-mitigation baseline and do not provide a direct comparison between different mitigation strategies. The probabilistic-loss-based cost-sensitive learning approach was not implemented for Random Forest and XGBoost models; therefore, the corresponding entries remain empty.

The results obtained using frequency-based binning reveal a strong dependence of mitigation performance on the underlying model type. For MLP models, all sampling-based methods generally perform better as pure over-samplers (without under-sampling), indicating that the removal of majority samples is detrimental in this setting. In contrast, Random Forest models often benefit from under-sampling. Across both model types, cSMOGN, crbSMOGN, and WERCS achieve some of the largest improvements in the very rare sample bin relative to the no-mitigation baseline. For XGBoost models, cost-sensitive learning yields the largest number of improvements, whereas among the sampling-based approaches the variants including under-sampling generally perform better.

The relevance-based evaluation largely confirms these observations. MLP models again show a clear preference for over-sampling without under-sampling, as all under-sampled variants perform worse than their corresponding over-sampling-only versions. Random Forest models exhibit a similar tendency toward under-sampling, although individual exceptions exist, most notably for crbSMOGN. For XGBoost models, DenseLoss again achieves the strongest improvements relative to the no-mitigation baseline.

One notable difference between the two evaluation schemes concerns crbSMOGN. While the method ranks among the strongest approaches under frequency-based binning, its performance appears less favorable when evaluated using relevance-based bins. A possible explanation is the mismatch between the relevance function used during mitigation and the one used for evaluation. Specifically, crbSMOGN relies on density-ratio relevance, whereas the relevance-based evaluation employs density-distance relevance. The resulting bin assignments may differ substantially, as density-ratio relevance directly quantifies the degree of imbalance as a ratio. All other relevance functions produce relevance scores bound between 0 and 1 and with different interpretations. Nevertheless, the present analysis alone does not allow conclusions regarding the superiority of one mitigation strategy over another, as only comparisons against the no-mitigation baseline are performed.

Overall, the full bin-win plots further emphasize the strong interaction between mitigation strategy and model type. The effectiveness of a given mitigation approach depends not only on the chosen relevance function and sampling strategy but also on the learning algorithm to which it is applied.

\end{document}